\documentclass[journal]{IEEEtran}

\ifCLASSINFOpdf
\else
\fi

\usepackage{amsmath}

\usepackage{booktabs}

\usepackage{multirow} 
\usepackage{graphics}
\usepackage[draft]{graphicx}
\usepackage{subfigure}
\usepackage{subcaption}
\usepackage[table]{xcolor}
\usepackage{ragged2e} 
\usepackage{tabularx}
\usepackage{CJK}
\usepackage{tikz}
\usepackage{cite} 
\usepackage[hidelinks]{hyperref}
\usepackage[edges]{forest}
\usepackage{amssymb}
\usepackage{colortbl,xcolor}

\usepackage{tikz}
\usetikzlibrary{trees, positioning, shapes.geometric} 

\begin{document}
%
\title{Image-to-Video Transfer Learning based on Image-Language Foundation Models: \\ A Comprehensive Survey}
%
%
%

\author{Jinxuan Li$^\dagger$, 
        Chaolei Tan$^\dagger$,
        Haoxuan Chen$^\dagger$,
        Jianxin Ma$^\dagger$,
        Jian-Fang Hu,
        Jianhuang Lai,
        Wei-Shi Zheng
\thanks{$\dagger$ means these authors contributed equally to this work.}
\thanks{Jinxuan Li, Chaolei Tan, Haoxuan Chen, Jianxin Ma, Jian-Fang Hu, Jianhuang Lai, and Wei-Shi Zheng are with the School of Computer Science and Engineering, Sun Yat-sen University. E-mail: \{lijx267, tanchlei, chenhx253,  majx33\}@mail2.sysu.edu.cn; \{hujf5, stsljh\}@mail.sysu.edu.cn; wszheng@ieee.org} 
}

%
%

\markboth{Journal of \LaTeX\ Class Files,~Vol.~14, No.~8, August~2015}%
{Shell \MakeLowercase{\textit{et al.}}: Bare Demo of IEEEtran.cls for IEEE Journals}
%



\maketitle

\begin{abstract}
Image-Language Foundation Models (ILFMs) have demonstrated remarkable success in vision-language understanding, providing transferable multimodal representations that generalize across diverse downstream image-based tasks. The advancement of video-text research has spurred growing interest in extending image-based models to the video domain. This paradigm, termed as image-to-video transfer learning, effectively mitigates the substantial data and computational demands compared to training video-language models from scratch while achieves comparable or even stronger model performance. This survey provides the first comprehensive review of this emerging field, which begins by summarizing the widely used ILFMs and their capabilities. We then systematically classify existing image-to-video transfer learning techniques into two broad root categories (frozen features and adapted features), along with numerous fine-grained subcategories, based on the paradigm for transferring image understanding capability to video tasks. Building upon the task-specific nature of image-to-video transfer, this survey methodically elaborates these strategies and details their applications across a spectrum of video-text learning tasks, ranging from fine-grained settings (e.g., spatio-temporal video grounding) to coarse-grained ones (e.g., video question answering). We further present a detailed experimental analysis to investigate the efficacy of different image-to-video transfer learning paradigms on a range of downstream video understanding tasks. Finally, we identify prevailing challenges and highlight promising directions for future research. By offering a comprehensive and structured overview, this survey aims to establish a structured roadmap for advancing video-text learning based on existing ILFM, and to inspire future research directions in this rapidly evolving domain. Additional resources and updates are maintained at our real-time github repository: \url{https://github.com/YuriPreisdent/awesome-image-to-video-transfer}.

\end{abstract}


%
\IEEEpeerreviewmaketitle

\section{Introduction}
%
%
%
%
\IEEEPARstart{M}{ultimodal} learning, particularly vision-language (or image-text) learning, has attracted significant attention in the community in recent years. With larger datasets and more powerful computing resources, researchers can train ILFMs to adapt to various downstream tasks and modalities. Vision-language models such as CLIP \cite{intro:clip} and BLIP \cite{intro:blip} demonstrate that large-scale pre-training on image-text pairs yields highly transferable visual representations. Subsequent approaches including MDETR \cite{intro:mdetr} and Grounding-DINO (G-DINO) \cite{intro:groundingdino} advance this paradigm by explicitly aligning free-form linguistic expressions with localized image regions, thereby bridging the gap between open-vocabulary language and visual grounding. More recently, LLM (Large Language Model)-based ILFMs\cite{llava,intro:qwenvl,intro:internvl} have emerged, which integrate diverse image-text understanding capabilities and can generate contextually appropriate responses in accordance with human instructions. Given that a video is fundamentally a sequence of images, a key research direction involves transferring the capabilities of ILFMs, which have demonstrated effectiveness on static image understanding, to the more complex video domain for solving challenging video understanding problems.

\begin{figure}[t!]
    \centering 
    \includegraphics[width=\linewidth]{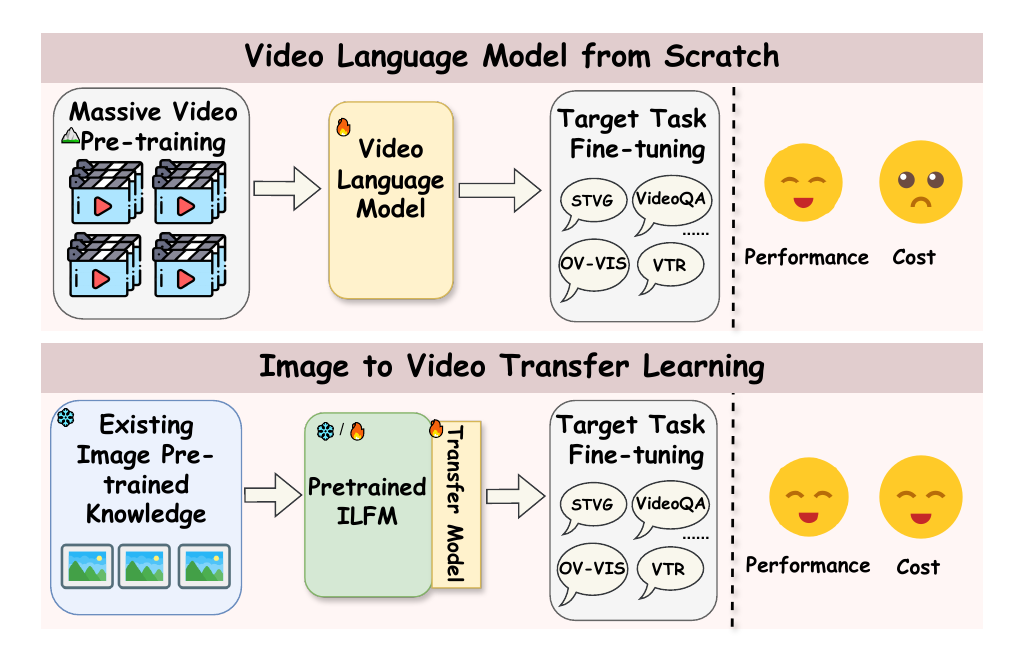}
    \caption{Comparison of two representative strategies for addressing video-text understanding problem with image-text and video-text foundation models. Pretraining video-based foundation models is much more challenging or even impractical, as they often contain more parameters and require larger amounts of training data and computational resources.}
    \label{intro}
\end{figure}

\begin{figure*}[t!]
    \centering 
    \includegraphics[width=\linewidth, height=0.35\linewidth]{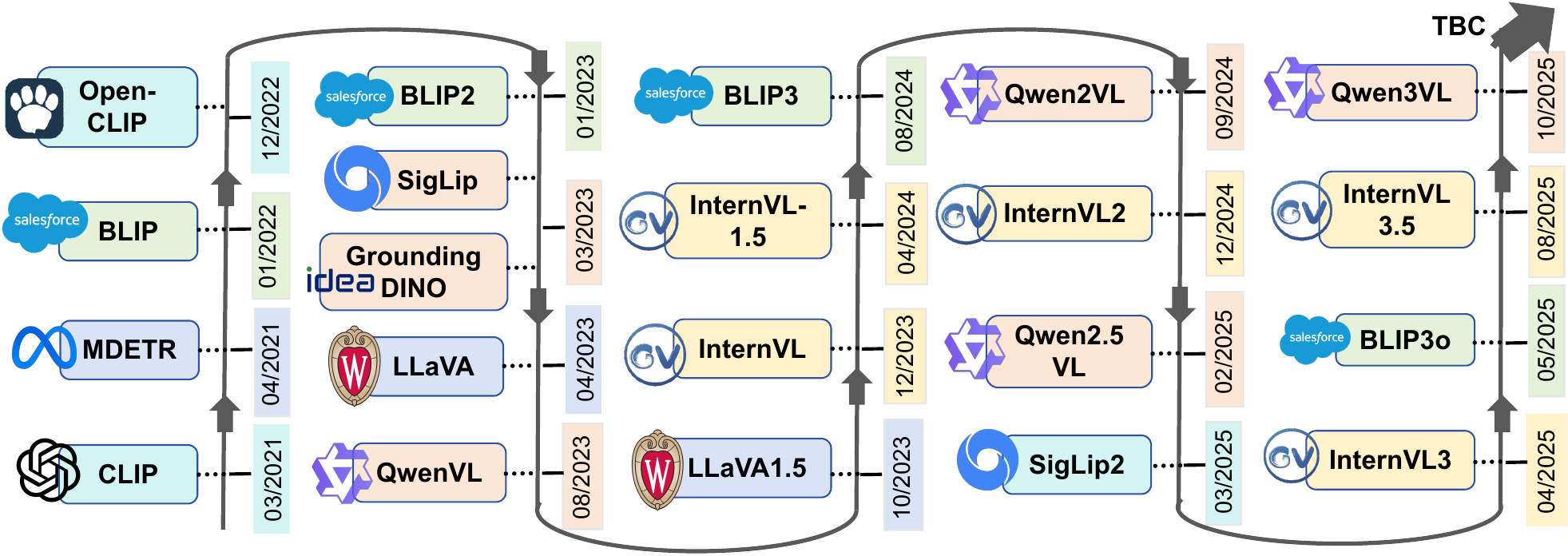}
    \caption{A comprehensive timeline depicting the development of image-language foundation models over recent years.
    }
    \label{timeline}
\end{figure*}
Compared to directly transferring models pre-trained solely on large-scale video-text data, adapting ILFMs to downstream video tasks demonstrates greater advantages. As illustrated in Fig.~\ref{intro}, training video-language foundation models from scratch encounters key challenges such as low computational efficiency and high data collection costs, since acquiring high-quality video-text data is inherently extremely difficult. In contrast, initializing from large-scale image-text pre-trained models presents higher practicality and scalability, owing to the richer availability and lower cost of image-text data. For instance, UniFormerV1 \cite{intro:uniformer} was pre-trained directly on the Kinetics-400 \cite{dataset:kinetics400} dataset, whereas UniFormerV2 \cite{var:uniformerv2} underwent post-pre-training starting from an image-text model. Compared to V1, UniFormerV2 achieves superior performance across diverse video understanding tasks with reduced training overhead, demonstrating that adapting image representations aligned with text to various video scenarios via task-specific transfer techniques is both more flexible and cost-effective.

The primary challenge in extending ILFMs to the video domain stems from the effective modeling of temporal information. In contrast to static images, videos comprise sequences of frames that encode complex spatiotemporal dynamics, including motion, event progression, and causal interactions. Consequently, an architectural transferring paradigm is necessary, moving from processing static spatial information to effectively modeling these intricate spatial-temporal relationships. Early methods processed each frame with 2D image networks and then applied a simple temporal fusion layer, which cannot effectively capture fine motion cues. Recently, researchers have been adapting these models to the video domain by inserting lightweight temporal modules into a frozen backbone and applying parameter-efficient fine-tuning techniques like adapters, often employing these strategies in hybrid configurations. Furthermore, a growing number of diverse approaches are continuously being introduced and explored in this domain. Side-tuning approaches~\cite{overview:side-tuning, intro:lst}, for example, is becoming increasingly popular as one of the most promising ways to accommodate the critical challenge of image-to-video transfer learning brought by the huge parameter and computation overheads. Other methods like distillation, also shows further potential to adapt the image-text aligned knowledge into the video domain especially under the era of Multimodal Large Language Models (MLLMs) in which we can capitalize on the exceptionally strong multimodal knowledge contained by these excessively large image-based MLLMs.


Despite the existence of numerous paradigms for image-to-video transfer learning, it remains underexplored which paradigm or combination thereof is most suitable for specific downstream tasks, as different tasks require models to possess distinct comprehension capabilities. For example, Spatio-Temporal Video Grounding (STVG) requires a model to enhance its capability for fine-grained temporal event localization while preserving its existing static spatial grounding performance. This objective is well-suited for the paradigm of fine-tuning with extra video models. In contrast, Video Question Answering (VideoQA) necessitates only a coarse-grained comprehension of video events but demands a more robust capacity for multi-modal knowledge understanding. Up to now, the optimal transfer learning paradigm for this latter task still remains an open question in the community.

In this survey, we provide a comprehensive overview of established strategies for migrating ILFMs to video-centric tasks. Beyond summarizing these techniques, we offer a critical analysis of their suitability for different application scenarios, concluding with a discussion on the prevailing challenges and promising directions for future research. The rest of this survey is organized as follows. Section II introduces the commonly used ILFMs developed in the community. Section III reviews and categorizes the techniques of leveraging image-level knowledge from ILFMs for video-language understanding. Sections IV and V systemically review the image-to-video transfer learning approaches for addressing varied downstream video-text understanding tasks, including fine-grained and coarse-grained problems, respectively. In Section VI, we provide the detailed experimental analysis. Finally, we provide promising image-to-video transfer learning research directions in Section VII and conclude in Section VIII.

\section{Overview of Image-Language \\ Foundation Models}
\begin{table*}[ht!]
\centering
\caption{Categorization and Analysis of popular Image-Language Foundational Models (ILFMs)}
\label{tab:foundation_models_simplified}
\footnotesize 
\newcolumntype{L}{>{\RaggedRight\arraybackslash}X} 
\begin{tabularx}{\textwidth}{@{} l p{3.2cm} p{3.5cm} p{3cm} p{4cm}  @{}}
\toprule
\textbf{Foundation Model} 
& \textbf{Pre-training Objective(s)} & \textbf{Pre-training Data} & 
\textbf{Primary Output} & \textbf{Key Capability} \\
\midrule

\textbf{CLIP~\cite{intro:clip}} & 
Image-Text Contrastive Learning & 
400M image-text pairs& 
Aligned Image \& Text Embeddings & 
Zero-Shot Image Classification, Robust Image-text Representation \\
\addlinespace

\textbf{MDETR~\cite{intro:mdetr}} & 
Phrase-to-Box Grounding & 
1.3M images with 11.5M phrase-box annotations & 
Bounding Boxes, Class Labels & 
Open-Vocabulary Object Detection \& Grounding \\
\addlinespace

\textbf{G-DINO~\cite{intro:groundingdino}} & 
Open-Set Object Detection & 
Over 3M images with 30M+ box annotations and 5M+ phrase relations& 
Bounding Boxes, Class Labels & 
Open-Vocabulary Object Detection \& Grounding \\
\addlinespace

\textbf{LLaVA~\cite{llava}} & 
Instruction Tuning on Image-Text Data & 
158K image-text instruction samples & 
Textual Response & 
Conversational VQA, Visual Reasoning \\
\addlinespace

\textbf{InternVL~\cite{intro:internvl}} & 
Image–Text Pre-training, Generative Training, Instruction Tuning &4B+ image–text pairs \& 100M+ curation data \& 1.5M+ instruction data 
& Aligned Vision–Language Embeddings, Textual Responses & 
Generic Visual Recognition, Image Retrieval, Captioning, VQA, Multi-Modal Dialogue \\
\addlinespace

\textbf{QwenVL~\cite{intro:qwenvl}} & 
Image–Text Pretraining, Multi-Task Learning, Instruction Tuning& 
1.4B+ image-text pairs \& Curated Billion-scale Bilingual Corpus & 
Textual Responses, Localized Bounding Boxes & 
Image Captioning, VQA, OCR-based Text Reading, Visual Grounding, Multilingual Dialogue \\
\bottomrule
\end{tabularx}
\end{table*}

Pretrained ILFMs have revolutionized a broad spectrum of visual understanding tasks such as object detection~\cite{rcnn,fastrcnn,fasterrcnn}, image captioning~\cite{showandtell,imagesemantic,densecap}, visual question answering~\cite{qavqa,NlcoVQA,simvqa}, etc. In this section, our goal is to introduce several representative ILFMs: CLIP~\cite{intro:clip} establishes image-text alignment through contrastive pretraining; MDETR~\cite{intro:mdetr} and G-DINO~\cite{intro:groundingdino} advance open-vocabulary detection and referring comprehension using language-aware visual encoding; lately LLaVA~\cite{llava}, InternVL~\cite{intro:internvl} and Qwen-VL~\cite{intro:qwenvl} integrate visual encoders with capable LLMs for instruction-based multimodal understanding. Collectively, these models bridge the gap between vision and language, providing powerful and adaptable foundations for various downstream vision-language applications.

\begin{figure*}[t!]
    \centering 
    \includegraphics[width=\linewidth,height=0.57\linewidth]{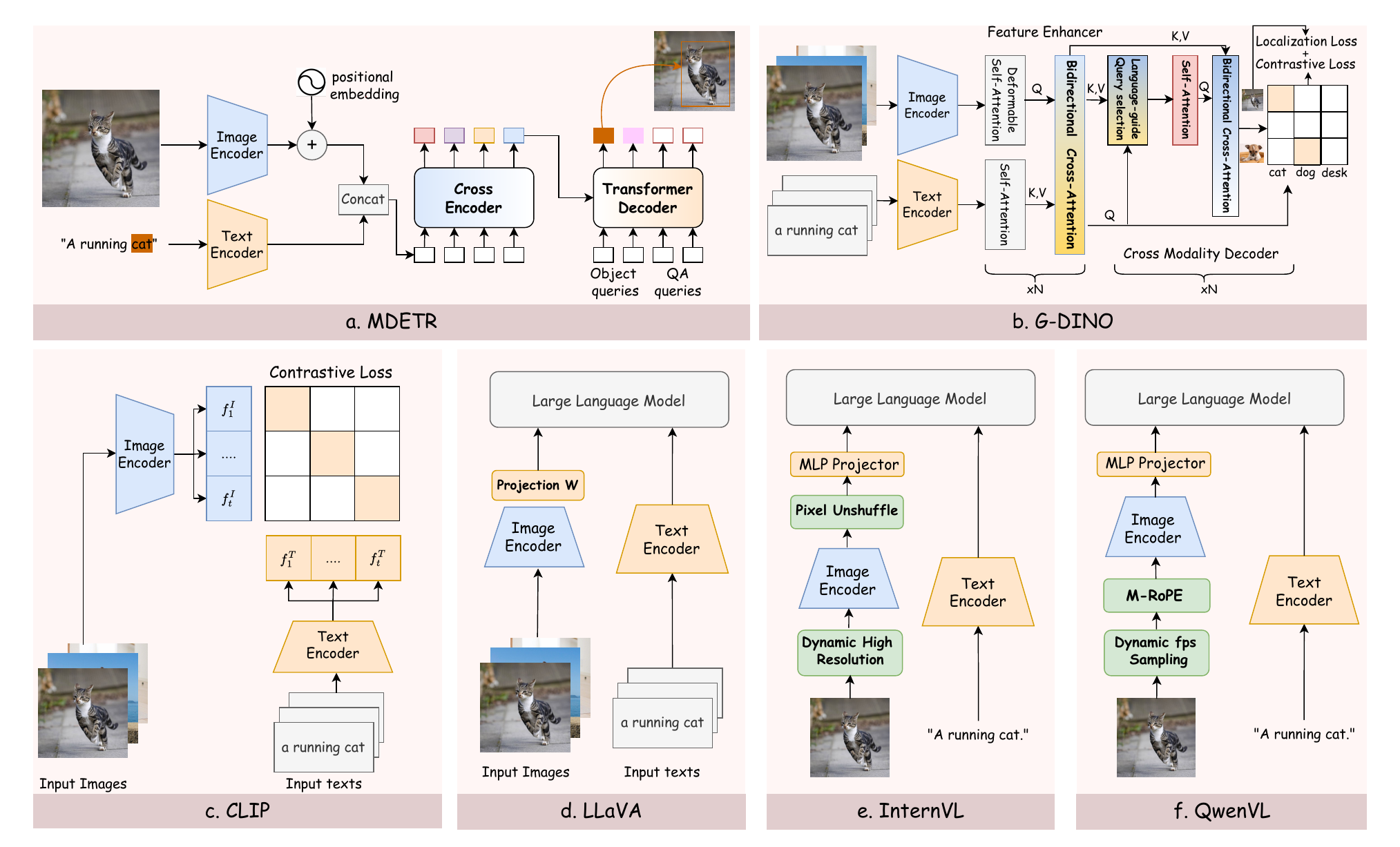}
    \caption{Popular ILFMs for image-to-video transfer learning in video-language understanding tasks.}
    \label{foundation model}
\end{figure*}

MDETR~\cite{intro:mdetr} is a representative foundation model for end-to-end grounded multimodal understanding, which can detect object locations in accordance with their textual descriptions in the given language query. As illustrated in Figure~\ref{foundation model}(a), it first employs a convolutional visual encoder (e.g., ResNet-101~\cite{resnet}) and a transformer textual encoder~\cite{roberta} to project both images and texts into a joint embedding space. Then a cross-encoder and a transformer decoder are utilized in cascade to obtain a grounded multimodal representation, on top of which a fine-grained contrastive loss aligns words or phrases in the text with their matching objects in the image. Learning such a grounded representation enables MDETR to reason over spatial relations and semantic bindings across modalities. Furthermore, MDETR is pre-trained on a massive data scale of $\mathord{\sim}$1.3M region-phrase pairs, which ensures robust learning of dense grounding between image objects and text tokens. Built upon the large-scale pretraining, MDETR exhibits strong generalization performance across various fine-grained tasks such as phrase grounding and referring expression comprehension.

G-DINO~\cite{intro:groundingdino} aims to localize and identify objects based on free-form textual descriptions. As illustrated in Figure \ref{foundation model}(b), the model builds upon a transformer-based architecture that integrates visual and linguistic modalities through a dedicated fusion module, enabling fine-grained cross-modal alignment between image regions and text queries. This design allows the model to effectively ground natural language phrases in images. The model is trained on a large-scale multi-source dataset comprising image-text pairs with bounding box annotations, which enables G-DINO to localize objects in unseen images using only text prompts—without any task-specific retraining, showcasing remarkable zero-shot capability. This strength makes it particularly suitable for applications in dynamic visual environments and open-ended retrieval tasks.

CLIP~\cite{intro:clip} is one of the earliest and most popular ILFMs that learns aligned representations of images and texts via large-scale contrastive pretraining. As illustrated in Figure~\ref{foundation model}(c), CLIP employs a dual encoder framework: a vision encoder built on scalable architectures such as vision transformers or convolutional networks, and a text encoder based on transformers. These encoders are trained to map paired images and texts into a shared latent space, where semantic alignment is achieved by maximizing similarity scores for relevant image-text pairs and minimizing them for irrelevant ones. A large-scale dataset consisting of $\mathord{\sim}$400M image-text pairs is collected from the Internet to train CLIP, enabling general vision-language alignment. Building upon this, subsequent research has focused on developing improved training recipes for CLIP-like pretraining for better scalability and training stability. Notably, OpenCLIP~\cite{intro:openclip} introduces an open-source framework to reproduce and extend the training of CLIP model. SigLIP~\cite{intro:siglip} is proposed to replace the standard InfoNCE objective with a simpler pairwise sigmoid loss. More recently, SigLIP2~\cite{intro:siglip2} makes further extensions by additionally including captioning-based objectives and self-supervised losses.

\definecolor{hidden-red}{RGB}{205, 44, 36}
\definecolor{hidden-blue}{RGB}{194,232,247}
\definecolor{hidden-orange}{RGB}{243,202,120}
\definecolor{hidden-green}{RGB}{34,139,34}
\definecolor{hidden-pink}{RGB}{255,245,247}
\definecolor{hidden-black}{RGB}{20,68,106}
\definecolor{purple}{RGB}{144,153,196}
\definecolor{yellow}{RGB}{255,228,123}
\definecolor{hidden-yellow}{RGB}{255,248,203}
\definecolor{tkcolor}{RGB}{224,223,255}
\definecolor{darkblue}{rgb}{0, 0.40, 0.75}

\tikzstyle{my-box}=[
rectangle,
draw=hidden-black, 
rounded corners,
text opacity=1,
minimum height=1.5em,
minimum width=5em,
inner sep=2pt,
align=center,
fill opacity=.5,
]
\tikzstyle{leaf3}=[
my-box,
minimum height=1.5em,
fill=yellow!20,
text=black,
align=left,
font=\normalsize,
inner xsep=5pt,
inner ysep=4pt,
align=left,
text width=45em,
]
\tikzstyle{leaf6}=[
my-box,
minimum height=1.5em,
fill=purple!30,
text=black,
align=left,
font=\normalsize,
inner xsep=5pt,
inner ysep=4pt,
]
\tikzstyle{leaf4}=[
my-box,
minimum height=1.5em,
fill=hidden-blue!57,
text=black,
align=left,
font=\normalsize,
inner xsep=5pt,
inner ysep=4pt,
]
\tikzstyle{leaf2}=[
my-box,
minimum height=1.5em,
fill=hidden-green!20,
text=black,
align=left,
font=\normalsize,
inner xsep=5pt,
inner ysep=4pt,
]
\tikzstyle{leaf}=[
my-box,
minimum height=1.5em,
fill=hidden-red!10,
text=black,
align=left,
font=\normalsize,
inner xsep=5pt,
inner ysep=4pt,
]
\tikzstyle{leaf5}=[
my-box,
minimum height=1.5em,
fill=darkblue!15,
text=black,
align=left,
font=\normalsize,
inner xsep=5pt,
inner ysep=4pt,
]

\begin{figure*}[!t]
\vspace{-2mm}
\centering
\resizebox{1\textwidth}{!}{
\begin{forest}
    forked edges,
    for tree={
    grow=east,
    reversed=true,
    anchor=base west,
    parent anchor=east,
    child anchor=west,
    base=left,
    font=\large,
    rectangle,
    draw=hidden-black,
    rounded corners,
    align=left,
    minimum width=4em,
    edge+={darkgray, line width=1pt},
    s sep=3pt,
    inner xsep=2pt,
    inner ysep=4pt,
    line width=1.1pt,
    ver/.style={rotate=90, child anchor=north, parent anchor=south, anchor=center},
    },
    where level=1{text width=11.5em,font=\normalsize,}{},
    where level=2{text width=13.7em,font=\normalsize,}{},
    where level=3{text width=14em,font=\normalsize,}{},
    where level=4{text width=50em,font=\normalsize,}{},
    [\textbf{Image-to-Video Transfer Learning}, ver, 
    [\textbf{\ \ Frozen Features}~(\S\ref{frozen})\qquad\ \ ,ver
    [\quad\textbf{\quad Knowledge Distillation}\quad\\
    \qquad\qquad\ \ \ (\S\ref{kd})\qquad\qquad
    [
        \textbf{Fine-grained:}
        OVTrack~\cite{OVMOT:OVTrack}{,} OVTrack+~\cite{OVMOT:OVTrack+}{,} VOVTrack~\cite{OVMOT:VOVTrack}{,} OVSORT~\cite{OVMOT:OVSORT}{,} 
        SLAck~\cite{OVMOT:SLATrack}{,} GLATrack~\cite{OVMOT:GLATrack}{,} \\
        UniVTG~\cite{tvg:univtg}{,}
        \\
        \textbf{Coarse-grained:}
        VideoDistill~\cite{vqa:videodistill}{,}
        CLIP2TV~\cite{vtr:clip2tv}{,}
        Text4Vis~\cite{var:text4vis}{,}
        MS-SL~\cite{vtr:prvr-conf}{,}
        FitCLIP~\cite{gvu:fitclip}{,}  \\
        CLIPPING~\cite{vtr:clipping}{,}
        TeachCLIP~\cite{vtr:teachclip}{,}
        MS-SL++~\cite{vtr:prvr-jour}{.}
        , leaf, text width=44em
    ]
    ]
    [\ \ \ \textbf{\quad \ Post-Network Tuning}\qquad\ \\
    \qquad\qquad\ \ \ (\S\ref{pnt})\qquad\qquad
    [            
        \textbf{Fine-grained:}
        NetTrack~\cite{OVMOT:NetTrack}{,}
        OVTR~\cite{OVMOT:OVTR}{,}
        LaMOTer~\cite{RMOT:LaMOT}{,}
        OV2Seg~\cite{OVVIS:OV2Seg}{,}
        UnLoc~\cite{tvg:univtg}{,}
        CLIP-VIS~\cite{OVVIS:CLIP-VIS}{,}
        \\
        MomentDETR~\cite{tvg:qvhighlight}{,}
        BAM-DETR~\cite{tvg:bamdetr}{,}
        UniVTG~\cite{tvg:univtg}{,}
        UnLoc~\cite{tvg:unloc}{,}
        TFVTG~\cite{tvg:tfvtg}{,}
        CoSPaL~\cite{stvg:CoSPaL}{,}
        STPro~\cite{stvg:STPro}{,} \\
        BriVIS~\cite{OVVIS:BriVIS}{,}
        OVFormer~\cite{OVVIS:OVFormer}{,}
        SpaceVLLM~\cite{stvg:spacevllm}{,}
        VideoTG-R1~\cite{VTG:VideoTGR1}{,}
        DVIS++~\cite{OVVIS:Dvis++}{.}
        \\
        \textbf{Coarse-grained:}
        BIMBA~\cite{vqa:bimba}{,}
        LinVT~\cite{vqa:linvt}{,}
        CLIP4Caption~\cite{caption:clip4caption}{,}
        E$^2$DVC~\cite{caption:e2dvc}{,}
        HierarQ~\cite{hierarq}{,}
        LeAdQA~\cite{vqa:leadqa}{,} \\
        UniFormerV2~\cite{var:uniformerv2}{,}
        SG-VLM~\cite{vqa:sgvlm}{,}
        OHRC~\cite{caption:ohrc}{,}
        TubeRMC~\cite{li2025tubermc}{.}
        , leaf, text width=44em] 
    ]
    [\qquad\quad\ \textbf{\quad Side-Tuning}\qquad\ \\
    \qquad\qquad\ \ \ (\S\ref{st})\qquad\qquad
    [            
        \textbf{Fine-grained:}
        R$^2$-tuning~\cite{tvg:r2tuning}{,}
        VDI~\cite{tvg:vdi}{,}
        PATF~\cite{tvg:patf}{.}
        \\
        \textbf{Coarse-grained:}
        CLIP2Video~\cite{vtr:clip2video}{,}
        STAN~\cite{gvu:stan}{,}
        EVL~\cite{var:evl}{,}
        DiST~\cite{var:dist}{,}
        MoTED~\cite{var:moted}{,}
        MAMS~\cite{caption:mams}{.}
        , leaf, text width=44em]
    ]
    ]
    [\textbf{Adapted Features}~(\S\ref{mf})\qquad\ \ , ver
    [\quad\textbf{\qquad \ Full Fine-Tuning}\quad\\
    \qquad\qquad\ \ \ (\S\ref{sf})\qquad\qquad
        [            
        \textbf{Fine-grained:}
        STVGBERT~\cite{stvg:STVGBert}{,}
        SWINBERT~\cite{caption:swinbert}{,}
        TubeDETR~\cite{stvg:TubeDETR}{,}
        STCAT~\cite{stvg:STCAT}{,}
        ReferDINO~\cite{RVOS:ReferDINO}{,}
        \\
        \textbf{Coarse-grained:}
        CLIP4Clip~\cite{vtr:clip4clip}{,}
        CLIP-Hitchhiker~\cite{vtr:clip-hitchhiker}{,}
        ViFi-CLIP~\cite{var:ViFi-CLIP}{,}
        TVR~\cite{vtr:disentangle}{,}
        CenterCLIP~\cite{vtr:centerclip}{,} \\
        X-CLIP~\cite{vtr:x-clip}{.}
        , leaf3, text width=44em]
    ]
    [\textbf{\qquad \quad\ \ Partial Tuning}\qquad\quad \\
    \qquad\qquad\ \ \ (\S\ref{partial-t})\qquad\qquad
        [
        \textbf{Coarse-grained:}
        Token-Mixing~\cite{vtr:tokmix}{,}
        TC-LLaVa~\cite{vqa:vcllava}{,}
        TEA~\cite{Zhang_Zeng_Shen_Wu_Zhou_Ma_2025}{,}
        mPLUG-2~\cite{caption:mplug2}{.}
        , leaf3, text width=44em]
    ]
    [\textbf{Fine-Tuning with Extra Models}\qquad \\
    \qquad\qquad\ \ \ (\S\ref{fef})\qquad\qquad
        [            
        \textbf{Fine-grained:}
        iKUN~\cite{RMOT:iKUN}{,}
        CoSD~\cite{stvg:CoSD}{,}
        CoSTA~\cite{stvg:CoSTA}{,}
        CGSTVG~\cite{stvg:CGSTVG}{,}
        TASTVG~\cite{stvg:TASTVG}{,}
        Vtimellm~\cite{tvg:vtimellm}{,} \\
        Seq2Time~\cite{tvg:seq2time}{,}
        NumPro~\cite{tvg:numpro}{,}
        Chrono~\cite{vtg:chrono}{,}
        NetTrack~\cite{OVMOT:NetTrack}{,}
        InstFormer~\cite{OVVIS:InstFormer}{,}
        Grounded-SAM-2~\cite{RVOS:GroundedSAM2}{,} \\
        UniTime~\cite{tvg:unitime}{,}
        AL-Ref-SAM-2~\cite{RVOS:ALRFSAM2}{.} \\
        \textbf{Coarse-grained:}
        VTG-LLM~\cite{caption:vtgllm}{,}
        X-Pool~\cite{vtr:x-pool}{,}
        MoMa~\cite{var:moma}{,}
        Vid2Seq~\cite{caption:vid2seq}{,}
        CM$^2$~\cite{caption:remember}{,}
        EvCap~\cite{caption:evcap}{,}
        \\
        BIKE~\cite{var:bike}{.}
        , leaf3, text width=44em]
    ]
    [\quad\textbf{\ Fine-Tuning with Adapter}\qquad\ \\
    \qquad\qquad\ \ \ (\S\ref{fa})\qquad\qquad
        [            
        \textbf{Fine-grained:}
        MASA~\cite{OVMOT:MASA}{,}
        LlaViLo~\cite{tvg:llavilo}{,}
        ReVisionLLM~\cite{tvg:revisionllm}{,}
        iKUN~\cite{RMOT:iKUN}{.}
        \\
        \textbf{Coarse-grained:}
        Tem-Adapter~\cite{vqa:temadapter}{,}
        ST-Adapter~\cite{var:st-adapter}{,}
        Adaptformer~\cite{var:adaptformer}{,}
        AIM~\cite{var:aim}{,}
        DUALPATH~\cite{var:dualpath}{,}
        \\
        TS2-Net~\cite{vtr:ts2net}{,}
        D$^2$ST~\cite{var:d2st}{,}
        MV-Adapter~\cite{vtr:mv-adapter}{,}
        Q-adapter~\cite{caption:qadapter}{,}
        ZeroI2V~\cite{var:zeroi2v}{,} \\
        TR-Adapter~\cite{vqa:tradapter}{.}
        , leaf3, text width=44em]
    ]
    [\textbf{\quad \ \ Fine-Tuning with LoRA}\qquad\ \\
    \qquad\qquad\ \ \ (\S\ref{fl})\qquad\qquad
        [            
        \textbf{Fine-grained:}
        ReferDINO~\cite{RVOS:ReferDINO}{,}
        VTG-LLM~\cite{caption:vtgllm}{,}
        VTimeLLM~\cite{tvg:vtimellm}{,}
        ReVisionLLM~\cite{tvg:revisionllm}{,}
        NumPro~\cite{tvg:numpro}{,}\\
        InstFormer~\cite{OVVIS:InstFormer}{.}
        \\
        \textbf{Coarse-grained:}
        DiscoVLA~\cite{vtr:discovla}{,}
        HierarQ~\cite{hierarq}{.}
        , leaf3, text width=44em]
    ]
    [\textbf{\qquad\quad\ \ Prompt Tuning}\qquad\quad \\
    \qquad\qquad\ \ \ (\S\ref{prompt-t})\qquad\qquad
       [            
        \textbf{Coarse-grained:}
        ATP~\cite{vqa:atp}{,}
        ActionCLIP~\cite{var:actionclip}{,}
        X-CLIP~\cite{var:x-clip}{,}
        ViTiS~\cite{vqa:zero}{,}
        Q-ViD~\cite{vqa:qvid}{,}
        VOP~\cite{vtr:vop}{,} \\
        CLIP-ViP~\cite{vtr:clip-vip}{.}
        , leaf3, text width=44em]
    ]
    ]
    ]
\end{forest}
}
\caption{{Taxonomy of image-to-video transfer learning paradigms and their corresponding methods.}
At top level, these methods can be grouped into two root paradigms: frozen features and adapted features, based on whether intermediate features extracted from the main information path of an ILFM remain totally unchanged or undergo adaptation. The frozen-features root paradigm includes paradigms of knowledge distillation, post-network tuning, and side-tuning, while the adapted-features root paradigm encompasses paradigms of full fine-tuning, partial tuning, fine-tuning with extra models, adapter, LoRA, and prompt tuning. For each specific transfer learning paradigm, we further divide its corresponding methods into fine-grained and coarse-grained types, reflecting different task requirements of the temporal and semantic alignment levels between visual and textual modalities.}
\label{fig:taxonomy-transfer-learning-methods}
\vspace{-1.0em}
\end{figure*}

LLaVA~\cite{llava} integrates visual understanding with natural language processing through an end-to-end trainable architecture. As Figure~\ref{foundation model} (d) shows, LLaVA consists of a pre-trained vision encoder CLIP-ViT and an LLM for multi-modal processing. It employs a simple projection layer to align visual representations with the LLM's embedding space, enabling cross-modality fusion. LLaVA undergoes two-stage training: first, feature alignment pretraining on $\mathord{\sim}$595K image-text pairs with a frozen vision encoder; second, instruction fine-tuning using $\mathord{\sim}$158K GPT-4-generated multi-modal instruction-response pairs to enhance reasoning and conversational capabilities. 
LLaVA-1.5 \cite{tvg:llava} extends LLaVA by replacing its image encoder with CLIP-ViT-L and vision-language projection layer with MLPs, which achieve better multi-modality understanding. Moreover, the model also introduces a multi-patch partitioning strategy, which improves the efficiency of processing high-resolution images and facilitates dynamic-resolution inputs. LLaVA-NeXT~\cite{intro:llava-next} quadruples the input image resolution to 672×672 and supports flexible aspect ratios. Using optimized visual instruction tuning on $<$1M samples, it improves reasoning, OCR, and world knowledge, achieving SOTA results and surpassing Gemini Pro in some tasks while retaining scalability.

As illustrated in Figure~\ref{foundation model}(e), InternVL~\cite{intro:internvl} is a large-scale vision-language foundation model with a vision encoder—InternViT-6B and a language middleware called QLLaMA, initialized from a multilingual LLaMA and augmented with cross-attention layers and learnable queries, with a total of 8B parameters. These components are aligned via a three-stage training strategy: starting with contrastive pre-training on 4.98B filtered image-text pairs from sources; followed by generative training on 1.03 billion high-quality captions using a combination of Image-Text Contrastive, Image-Text Matching, and Image-grounded Text Generation losses; and concluding with supervised fine-tuning on 4 million instruction samples for multimodal dialogue. 
Several extensions have been developed. InternVL2~\cite{intro:internvl2.5} extends the framework to handle video and multimodal data. InternVL3~\cite{intro:internvl3} demonstrates improved GUI understanding and spatial reasoning capabilities through a single-stage native multimodal pre-training.

Qwen-VL~\cite{intro:qwenvl} is a vision-language foundation model trained via a multi-stage strategy (from pre-training to supervised fine-tuning) rather than simple visual encoder-LLM concatenation. As shown in Figure~\ref{foundation model}(f), its architecture includes a ViT-based visual encoder, a position-aware adapter, and the Qwen language model. The model is pre-trained on a massive dataset of over 1.4 billion image-text pairs and fine-tuned through a multi-stage training process, including large-scale weakly supervised and multi-task supervised fine-tuning. A key strength of Qwen-VL is its comprehensive multimodal capability, supporting multiple image inputs, multilingual conversations, and fine-grained recognition through its multi-scale training strategy and high-resolution processing (up to 448×448 pixels). Qwen-VL-Chat~\cite{intro:qwenvl} enhances Qwen-VL with sophisticated multi-image dialogue capabilities, supporting complex interleaved vision-language tasks through improved alignment training and safety mechanisms. Qwen2-VL~\cite{intro:qwen2vl} further enhances visual reasoning, context handling, and multilingual performance, while Qwen2.5-VL~\cite{intro:qwen2.5vl} achieves significant breakthroughs through its upgraded vision encoder with dynamic high-resolution support (up to 1536×1536). It demonstrates superior performance on complex tasks like chart and document analysis, fine-grained VQA, long-context multimodal dialogue, and complex instruction following, achieving new SOTA across multiple vision-language benchmarks.

\textbf{Discussion:} All the above ILFMs not only open up new opportunities for developing language-based image understanding models, but also set the stage for their further application in more complex video analysis tasks, where the integration of temporal information becomes essential. Among these ILFMs, MDETR and G-DINO are featured by their stronger abilities in capturing fine-grained visual–textual correspondences, since they are explicitly trained with grounding supervision to align object regions with textual phrases. In contrast, CLIP is more adept at handling coarse-grained visual–textual correspondences, due to its global-level contrastive pretraining objective. Besides, LLM-based models such as LLaVA, InternVL, and Qwen-VL empower ILFMs with strong instruction-following abilities, which is crucial for addressing more complex tasks such as reasoning and planning. Thus, they are expected to become mainstream for some advanced applications in future.

\section{Overview of Image-to-Video \\ Transfer Learning}
In this section, we aim to meticulously construct an extensible taxonomy, in which any existing image-to-video transfer learning method can be classified into at least one technical paradigm. It first categorizes transfer methods into frozen-features and adapted-features paradigms depending on whether the features extracted by ILFMs remain unchanged or undergo adaptation. More specific sub-categories under these two root categories will be elaborated in the following. It is noted that we focus on transfer learning~\cite{tlsur-1,tlsur-2,tlsur-3} involving parameter updates, while zero-shot learning~\cite{zssur-1,zssur-2,zssur-3} are out of scope.

\subsection{Frozen Features}
\label{frozen}

\begin{figure*}[t!]
    \centering
    \includegraphics[width=\linewidth]{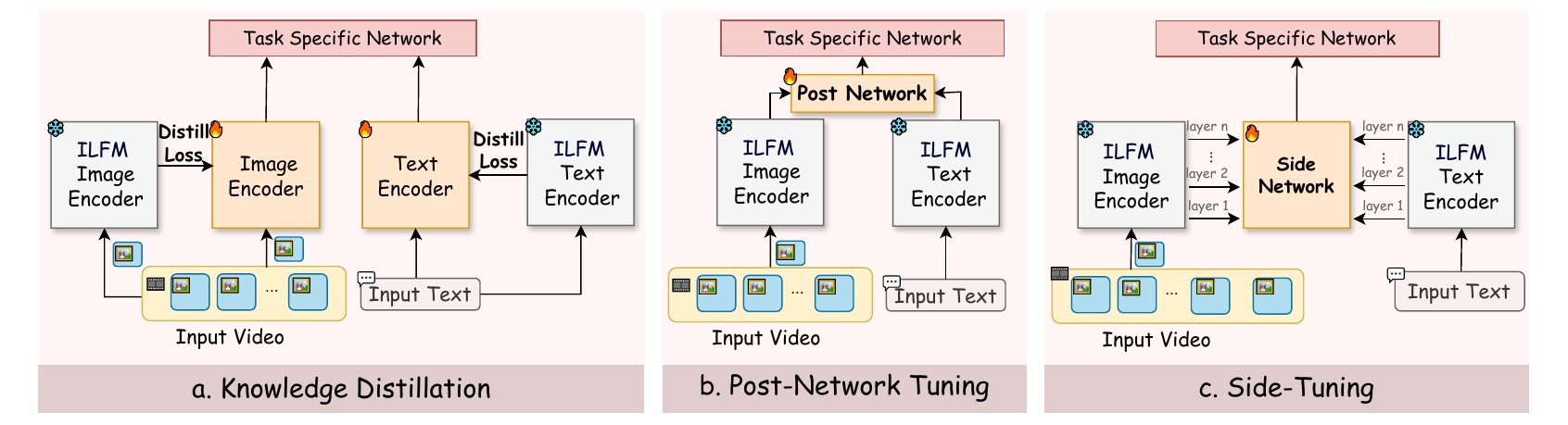}
    \caption{Methods and architectures of transferring pre-trained ILFM to video domain via frozen features. 
    }
    \label{transfer learning frozen}
\end{figure*}

By treating video as a sequence of image frames in temporal order, a considerable number of existing methods directly leverage pretrained ILFMs to extract frozen frame-level features, regarding them as guidance signal, initial or auxiliary inputs to conduct image-to-video transfer learning. Specifically, representative lines of research under such principle include knowledge distillation, post-network tuning, and side-tuning.

\subsubsection{Knowledge Distillation}
\label{kd} As shown in Figure~\ref{transfer learning frozen}(a), within the knowledge distillation paradigm, the common practice is to employ a pretrained ILFM as the teacher model to enhance the cross-modal alignment capabilities of the visual and textual encoders in the downstream video-text understanding models. Various distillation objectives can be defined to facilitate targeted image-to-video knowledge transfer between the teacher and student encoders, often working in conjunction with another downstream loss. For instance, the distillation paradigm has been widely adopted in early Open-Vocabulary Multi-Object Tracking (OV-MOT) methods~\cite{OVMOT:OVTrack,OVMOT:OVTrack+,OVMOT:VOVTrack,OVMOT:OVSORT} to inherit open-vocabulary capabilities from ILFMs like CLIP. For better distillation transfer performance, one critical design point is on how to effectively balance the learning of temporal modeling knowledge from downstream data and strong cross-modal alignment knowledge from the pretrained ILFMs.

\subsubsection{Post Network Tuning}
\label{pnt}
Many methods achieve temporal modeling by additionally introducing a post network to process the frozen features obtained for each sampled frame (shown in Figure~\ref{transfer learning frozen}(b)). One straightforward scheme here is to construct the post network with temporal attention mechanism, enabling the information exchange across varied frames. However, this paradigm is usually limited in performance, as it is hard to directly capture dynamic temporal patterns from static features. To more effectively harness the visual potential of frozen large-scale image-text models, researchers employ a Q-Former~\cite{intro:bilp2} architecture, utilizing a set of learnable queries to interact with features from all video frames. Through this approach, queries can dynamically capture temporal dependencies and motion patterns in videos, rather than supplying image features with extra temporal cues. This not only reduces computational costs but also unlocks the potential of ILFMs. Beyond general temporal modeling, many methods have incorporated more task-specific temporal modeling strategies, such as Hungarian algorithm or Brownian bridge-based association. In summary, this paradigm relies on high-quality image-text features and carefully designed, task-specific temporal modeling architectures to achieve strong video understanding performance.

\subsubsection{Side-Tuning}
\label{st}
as illustrated in Figure~\ref{transfer learning frozen}(c), Side-tuning~\cite{overview:side-tuning} often introduces a learnable side network in parallel to the main information paths of image/text encoders in the ILFMs. It receives the layer-wise latent features extracted by ILFM encoders as input, facilitating hierarchical spatial modeling and temporal refinement in a coarse-to-fine manner. In practice, the side network often adopts a lightweight design, incorporating functions like text-modulated spatial pooling and temporal modeling operations, etc. During training, gradients are effectively confined within the side-blocks, ensuring both parameter and memory efficiency. Recent studies~\cite{tvg:r2tuning,var:moted} show that the system can be further enhanced by connecting side-blocks with the outputs of CLIP, which achieves better results on several downstream video-text understanding tasks.

\textbf{Discussion:} Frozen-feature-based image-to-video transfer learning frameworks offer notable advantages in adapting image-based foundation models to video–language understanding tasks. They maintain high parameter efficiency by freezing pretrained architecture and weights, thereby mitigating catastrophic forgetting of previously acquired knowledge. However, their performance may be limited by the inherent representational capacity bottleneck of static image-level features and their imperfect alignment with video task requirements, particularly when generalizing to novel scenarios.

\subsection{Adapted Features}
\begin{figure*}[t!]
    \centering 
    \includegraphics[width=0.96\linewidth]{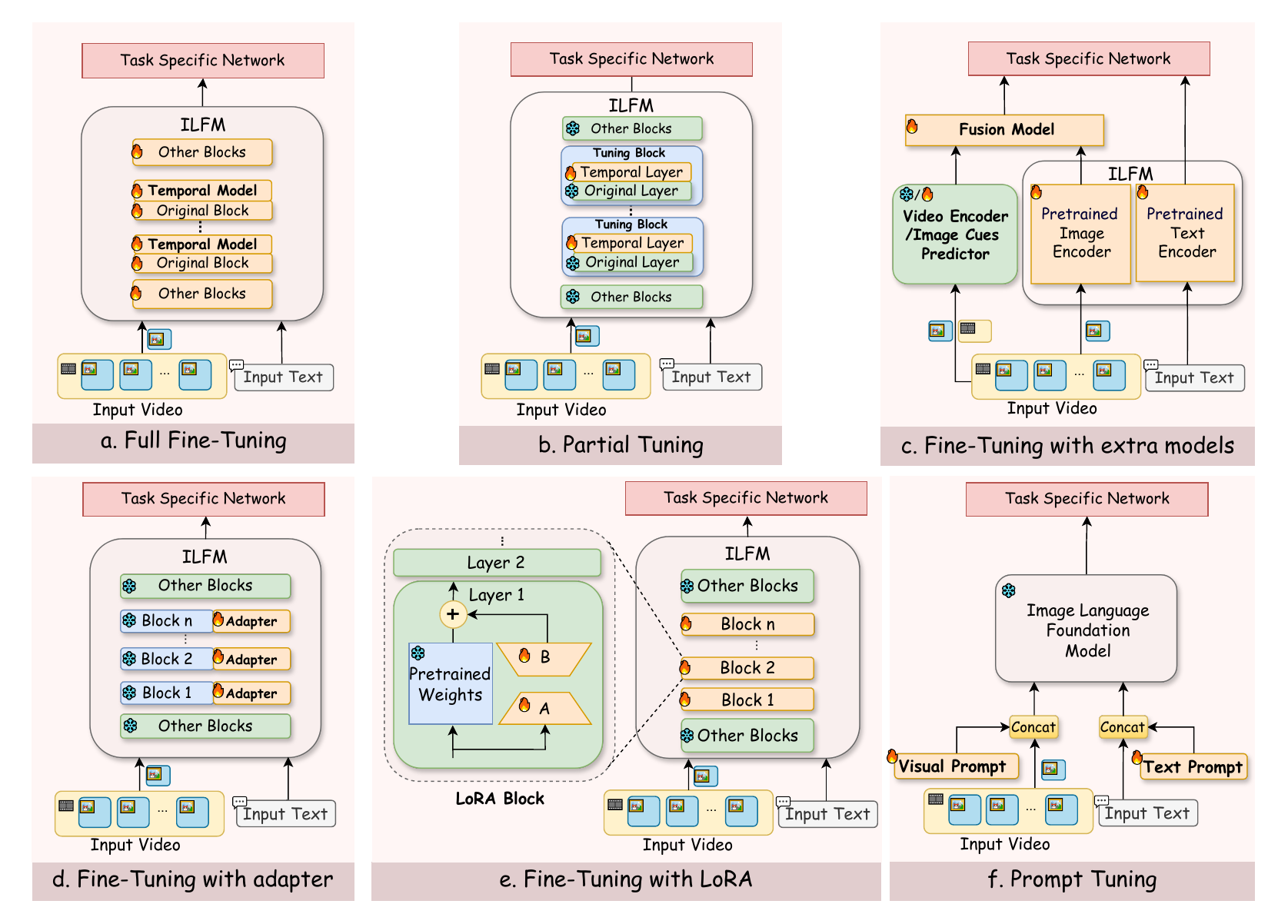}
    \caption{Methods and architectures of transferring pre-trained image-text model to video domain via adapted features. 
    }
    \label{transfer learning modified}
\end{figure*}

Instead of directly using frozen features, many approaches perform image-to-video transfer learning by explicitly adjusting the backbone model architecture and parameters, which we classify as adapted features. In the following, we will elaborate on the techniques, including Full Fine-Tuning, Partial Tuning, Fine-Tuning with Extra Models, Fine-Tuning with Adapter, Fine-Tuning with LoRA, and Prompt Tuning, whose architectures are presented in Figure \ref{transfer learning modified}, respectively.
\label{mf}

\subsubsection{Full Fine-Tuning}
\label{sf}
It is straightforward to empower a pre-trained ILFM with the ability of video-text understanding by building a temporal block on top of each block of ILFM and then simply fine-tune all the parameters involved in the temporal modeling components together with the ILFMs~\cite{stvg:STVGBert,stvg:TubeDETR,stvg:STCAT,caption:swinbert}. Although simple and effective, these approaches indeed are less applicable in practice as they need a large-scale, well-defined, and annotated data and a large amount of computing power in the model training stage.

\subsubsection{Partial Tuning}
\label{partial-t}
Partial tuning strikes a balance between full fine-tuning and more parameter-efficient methods like adapters or prompt tuning. This strategy involves updating only a specific subset of the pre-trained ILFM's parameters, while freezing the rest of the model parameters. Recently, several methods~\cite{EDIT:P2V,EDIT:TOKEN,EDIT:CNET} have introduced temporal attention modules into the original spatial attention modules, enabling ILFMs to acquire temporal–spatial modeling capability. Such tuning strategies strike a favorable balance between performance and computational cost, while still relying heavily on the inherent spatial modeling capacity of the pretrained ILFMs, as they do not involve full model training.

\subsubsection{Fine-Tuning with Extra Models}
\label{fef}
Owing to the limited capacity of image-level features in capturing dynamic information, numerous image-to-video transfer learning methods incorporate auxiliary models to enhance temporal modeling. Currently, two dominant paradigms exist. The first paradigm augments image-based features with dynamic signals (e.g., optical flow, 3D spatio-temporal features) and introduces specialized modules to fuse or align them.
For example, in STVG, recent methods employ both image and video backbones (e.g., SlowFast\cite{slowfast} or Video Swin Transformer\cite{VideoSwin}), coupled with attention-based fusion modules, for collaborative spatio-temporal modeling. 
The second paradigm leverages the structural outputs of external models—such as detection results, segmentation masks, and motion saliency maps—to guide image-to-video transfer with lower computational cost. In summary, fine-tuning with extra models offers an effective strategy for injecting video-specific structural information into ILFMs. It is particularly suitable when video annotations are limited or noisy, and has demonstrated strong empirical performance across tasks such as referring video object segmentation, temporal grounding, and video-language retrieval.

\subsubsection{Fine-Tuning with Adapter}
\label{fa}
Adapter-based methods offer a parameter-efficient solution to adapting ILFMs to video-language understanding tasks. This is achieved by inserting lightweight, trainable modules, called adapters, into intermediate layers of the frozen pre-trained models ~\cite{overview:trans,tvg:mul,tvg:vdi}. While adapters are a general-purpose mechanism, their specific design is task-dependent. For video understanding, a common and effective design is to construct the adapter as a stack of temporal attention and cross-frame interaction layers, which are crucial for modeling the temporal dynamics in video data.
By keeping the parameters frozen, it retains ILFMs' visual-semantic understanding capabilities while efficiently learns video-specific temporal patterns through adapters. 
Recently, this technique has been particularly attractive in scenarios with limited video annotations or computational budgets, and it emerges as a prominent direction for addressing downstream tasks like VideoQA, TVG, and video captioning.

\subsubsection{Fine-Tuning with LoRA}
\label{fl}
Low-Rank Adaptation (LoRA)~\cite{LoRA} provides an alternative paradigm for ILFM adaptation. Instead of directly learning to update model weights $\Delta W$, LoRA approximates it with two small, trainable low-rank matrices, i.e., $\Delta W = BA$, where $B \in \mathbb{R}^{d \times r}$, $A \in \mathbb{R}^{r \times k}$, and the rank $r \ll \min(d, k)$. 
This implicitly constrains the model's exploration within a low-dimensional subspace, effectively mitigating the risk of overfitting, which is particularly beneficial when labeled video data is scarce. Thus, LoRA is widely used to efficiently adapt large-scale foundation models to various downstream video tasks. However, as LoRA cannot model temporal information itself, it requires external temporal modules or networks, such as those used in full fine-tuning or post-network tuning.

\subsubsection{Prompt Tuning}
\label{prompt-t}
Prompt tuning has recently emerged as a highly parameter-efficient alternative for adapting ILFMs to video-related tasks~\cite{vqa:zero,vqa:qvid}. By freezing the backbone and introducing a small set of learnable prompts, the model is guided to capture temporal cues without modifying its core parameters. A common design is to share a single set of prompts across all frames to enforce temporal consistency, or to propagate prompts sequentially across frames so that information accumulates and evolves with motion. These temporal variants enable the frozen ILFMs to move beyond static frame modeling and capture motion cues and cross-frame dependencies. This strategy achieves efficient spatio-temporal adaptation with low computational cost, though its reliance on frozen backbones may restrict fine-grained temporal reasoning compared to full or partial fine-tuning.

{\textbf{Discussion:} The aforementioned image-to-video transfer learning methods primarily adapt existing image–text foundation models to address video–text understanding tasks through two main directions: enhancing the encoders of established video understanding models or incorporating additional learnable temporal modules into ILFMs. Notably, these approaches are not mutually exclusive. Existing studies have shown that their integration can yield complementary benefits and further improve transfer performance. For example, fine-tuning with auxiliary models can be seamlessly combined with parameter-efficient strategies such as LoRA, facilitating the development of more advanced and robust video understanding models.
}

\section{Fine-grained Video-Text Understanding}
In the following, we review methods that adapt existing pre-trained ILFMs to address specific video-text understanding tasks. Based on the required granularity of understanding, we categorize the video-text understanding tasks into fine-grained and coarse-grained classes, providing a clearer structural understanding and enabling researchers to identify methods most relevant to their research domains. In the following, we first introduce fine-grained video-text understanding, which requires precise localization of spatial regions (e.g., bounding boxes or segmentations) or temporal intervals. 

\subsection{Open Vocabulary Multi-Object Tracking}
Open Vocabulary Multi-Object Tracking (OV-MOT) 
aims to track all the objects described by input text in an open-vocabulary manner, rather than relying on a predefined, closed set of categories. It is a task extended from traditional MOT, in which the objects of interest are explicitly referred to with natural language descriptions. 

Most of existing OV-MOT methods \cite{OVMOT:MASA,OVMOT:OVTR,OVMOT:GLATrack,OVMOT:OVTrack+,OVMOT:OVTrack,OVMOT:NetTrack,OVMOT:SLATrack,OVMOT:VOVTrack} mainly follow a frozen feature paradigm to transfer useful knowledge from pre-trained CLIP and G-DINO models. 
For instance, OVTrack~\cite{OVMOT:OVTrack} and OVTrack+~\cite{OVMOT:OVTrack+} develop knowledge distillation blocks to enhance the RoI features extracted by ResNet50 and FPN with pre-trained CLIP, Based on OVTrack,  VOVTrack~\cite{OVMOT:VOVTrack} further introduce object appearance self-consistency and spatial-appearance mutual-consistency constraints to capture spatial relationships across frames, where object appearance is directly extracted with CLIP. SLAck~\cite{OVMOT:SLATrack} introduces a spatial-temporal object attention mechanism to jointly model semantic, location, and appearance cues, in which knowledge in CLIP is distilled to enhance the semantic head. GLATrack~\cite{OVMOT:GLATrack} employs a pre-trained CLIP model to extract semantic information of target objects from the reference frame, which is subsequently used to mitigate the impact of erroneous features in key frames. OVTR~\cite{OVMOT:OVTR} employs a post-network fine-tuning strategy, wherein a Transformer-based dual-branch decoder is introduced to process frozen features extracted by CLIP. These methods rely on coarse-grained image-text correspondences captured by CLIP, which restricts the model's ability to capture the open-world objects with high dynamicity. NetTrack~\cite{OVMOT:NetTrack} employs G-DINO~\cite{intro:groundingdino}) to establish fine-grained object-text correspondences and a point tracker (e.g., CoTracker~\cite{CoTracker}) to trace trajectories of points of interest, hence enhancing cross-frame object association. MASA~\cite{OVMOT:MASA} introduces a plug-and-play tracking adapter t
o enhance open-world foundation models (e.g., G-DINO) for tracking arbitrary detected objects.

\subsection{Temporal Video Grounding}
Temporal Video Grounding (TVG) is an important fine-grained multimodal video understanding task to localize the temporal boundaries of given language queries, which are typically in the form of single sentences~\cite{tvg:tall, tvg:didemo, tvg:qvhighlight}, or even more complicated formats like paragraphs~\cite{tvg:depnet, tvg:hscnet, tvg:siamgtr, tvg:synopground} or articles~\cite{tvg:wsag, tvg:mvtcl}. Early approaches in TVG predominantly relied on a two-stage and proposal-based pipeline, where a large number of predefined candidate temporal segments are generated and then ranked according to their relevance to the language query. Afterwards, the field has shifted towards methods which are in a one-stage pipeline and those that can leverage powerful pre-trained image foundation models like CLIP~\cite{intro:clip} via image-to-video transfer learning. A pivotal development was the introduction of DETR-style frameworks~\cite{intro:detr} that follow a Post-Network Tuning scheme. Moment-DETR~\cite{tvg:qvhighlight} pioneered this by reformulating TVG as a direct set-prediction problem, utilizing a trainable Transformer-based head over frozen backbone encoders. Building on this, BAM-DETR~\cite{tvg:bamdetr} introduced a boundary-oriented formulation and a dual-pathway decoder to achieve more precise temporal alignment. UniVTG~\cite{tvg:univtg} proposed a unified multi-modal grounding head to handle multiple tasks within the same framework. UnLoc~\cite{tvg:unloc} employed a post-pretraining scheme to fuse pyramid features from a CLIP model for both frames and text.

Beyond the post-network paradigm, other transfer strategies have also been explored. VDI~\cite{tvg:vdi} introduces a language-side adaptation paradigm in which the CLIP visual encoder remains frozen, while the text encoder is adapted by infusing video-derived supervisory signals through a side-tuning approach. R$^2$-Tuning~\cite{tvg:r2tuning} implements side-tuning using a reversed recurrent tuning block to explicitly perform query-modulated spatial pooling and coarse-to-fine temporal refinement. To incorporate explicit relational reasoning, PaTF~\cite{tvg:patf} encodes external cues by introducing a parallel transformer that processes structured scene graph representations.

More recently, Multimodal Large Language Models are introduced to enhance temporal reasoning. LLaViLo~\cite{tvg:llavilo} utilizes a video semantic modeling adapter to inject a joint feature representation into an LLM. To incorporate explicit time information as external cues, Chrono~\cite{vtg:chrono} introduced a sequence-to-sequence approach that integrates video frames with their corresponding timestamp cues. Similarly, VTG-LLM~\cite{caption:vtgllm} integrated timestamp knowledge through sequence-time embeddings, absolute-time tokens, and slot-based token compression. ReVisionLLM~\cite{tvg:revisionllm} incorporated a recursive processing architecture with hierarchical adapters to progressively refine temporal segments in hour-long videos. In a training-free manner, TFVTG~\cite{tvg:tfvtg} employed LLMs to reason about sub-event relationships while using separate VLMs to localize each sub-event. NumPro~\cite{tvg:numpro} employs a fine-tuning with extra models strategy to achieve video-language understanding. It overlays unique numerical identifiers onto each video frame, thereby translating the temporal grounding task into a visual OCR problem that can be directly resolved by a Video LLM.

\subsection{Spatio-Temporal Video Grounding}
STVG aims to predict a spatio-temporal tube for the event depicted in given language query from an untrimmed video. 
Earlier methods \cite{stvg:VidSTG, stvg:OMRN, stvg:HCSTVG} use unimodal detectors like Faster R-CNN~\cite{fasterrcnn} to detect objects in each video frame and then ground the temporal locations based on the detection features. STVGBert \cite{stvg:STVGBert} is the first work to introduce and fully finetune a pre-trained ViLBERT \cite{ViLBERT} model and develop a novel spatio-temporal composition/decomposition attention module, thereby enabling joint spatio-temporal modeling. Recent studies \cite{stvg:TubeDETR,stvg:STCAT,stvg:CoSD,stvg:sgfd,stvg:CoSTA,stvg:TASTVG,stvg:CGSTVG} have developed frameworks based on the powerful pre-trained MDETR model \cite{MDETR}, incorporating additional spatio-temporal interaction mechanisms. To achieve better performance, some approaches \cite{stvg:TubeDETR,stvg:STCAT} additionally introduce a temporal module to capture the cross-frame interaction and fine-tune all the parameters involved in the model. Specifically, TubeDETR \cite{stvg:TubeDETR} introduces a fast visual-only branch in the encoder and a time-aligned cross-attention mechanism in the decoder, enabling object queries to achieve joint spatio-temporal modeling. STCAT \cite{stvg:STCAT} designs a video-level multimodal template in the decoder to maintain temporal consistency across spatial predictions.

Beyond adapting image backbones for dynamic information modeling, recent SOTA methods further incorporate video-level features from additional video backbones to enhance the capture of temporal dynamics.
CoSD~\cite{stvg:CoSD} leverages SlowFast~\cite{slowfast} features and proposes a two-stream framework with cross-stream collaboration to capture static appearance and dynamic motion cues collaboratively for target localization. CoSTA~\cite{stvg:CoSTA} employs a static-dynamic encoder-decoder architecture, while additionally utilizing video features extracted by 3D-CNN~\cite{3dCNN}.  During the decoding process, it entangles temporal queries with spatial features and bridges temporal 
and spatial localizations. CGSTVG~\cite{stvg:CGSTVG} and TA-STVG~\cite{stvg:TASTVG} integrate VidSwin~\cite{VideoSwin} and MDETR to extract target-aware spatial and temporal cues under textual guidance.

\subsection{Video Segmentation}
Open Vocabulary Video Instance Segmentation (OV-VIS) extends traditional VIS by enabling the segmentation and tracking of of unseen object categories, leveraging vision-language foundation models to generalize beyond a predefined label set. Current research mainly leverages CLIP's open-vocabulary capabilities for this task, designing additional network modules after the CLIP encoder to enable image-to-video transfer.
OV2Seg~\cite{OVVIS:OV2Seg} is the first to integrate CLIP into a post-network fine-tuning strategy for OVVIS. It adopts a DETR-like architecture with object-centric queries for mask prediction and cross-frame matching. The query embeddings, which combine information from current and previous frames, are compared with CLIP-extracted text features for open-vocabulary classification. InstFormer~\cite{OVVIS:InstFormer} integrates instance-guided attention into CLIP's visual encoder, using segmentation cues to aggregate text-aligned features, which are then fed into the tracking module for instance matching across frames. BriVIS~\cite{OVVIS:BriVIS} uses a frozen video instance segmentor to generate frame-level queries and integrates multi-scale CLIP visual features for temporal modeling. It introduces a Brownian bridge to constrain instance features, using a contrastive objective to align the bridge center with the correct class text embedding while distancing it from irrelevant classes. OVFormer~\cite{OVVIS:OVFormer} constructs video-level instance queries and aligns them with frozen CLIP features using cross-attention for unified segmentation and tracking across frames. OV-DVIS++~\cite{OVVIS:Dvis++} propose a address the video segmentation problem via a segment-tracking-refinement procedure in a local to global manner. CLIP-VIS~\cite{OVVIS:CLIP-VIS} directly leverages CLIP’s image features to generate masks by combining predicted object scores and mask IoU. To alleviate the effect of historical noise, it first applies Hungarian algorithm to identify the $K$ most relevant frames from memory, and then performs instance tracking. 



\section{Coarse-grained Video Understanding}
Compared to fine-grained video-text understanding, coarse-grained tasks do not require precise temporal grounding results. They place greater emphasis on the holistic comprehension of events depicted in both text and video, such as determining whether a text-described event occurs in a video (Video QA) or generating textual descriptions of video content (Video Captioning). In this section, we review methods for coarse-grained video-text understanding tasks.

\subsection{Video-Text Retrieval}
Video–Text Retrieval (VTR) aims to retrieve the most relevant video from a large corpus given a textual query, or vice versa. Since the introduction of the large-scale image–text pretrained model CLIP~\cite{intro:clip}, a series of subsequent studies have focused on transferring its knowledge to video–text retrieval, achieving substantial performance gains over earlier methods. Early approaches in this domain typically adopt a full fine-tuning paradigm, where model parameters are initialized from pretrained weights and subsequently optimized for downstream tasks. CLIP4Clip~\cite{vtr:clip4clip} is the first method that explored transferring CLIP’s image representations to the video domain for video–text retrieval. It employs a 2D or 3D linear projection layer on flattened patches and introduces an additional learnable token to extract frame-level embeddings, following a pipeline similar to that of the vanilla ViT~\cite{vit}. In addition, a preliminary video-text post-pretraining approach was explored in CLIP4Clip~\cite{vtr:clip4clip} to further adapt CLIP representations to the video domain. CLIP2Video~\cite{vtr:clip2video} extends image–text pretrained models to video–text retrieval by introducing a temporal difference block and a temporal alignment block. The temporal difference block enhances motion-related cues by modeling differences between adjacent frames, while the temporal alignment block aggregates text and frame embeddings through learnable center embeddings in the joint space. Another method~\cite{vtr:disentangle} improves the full fine-tuning performance by developing a weighted token-wise interaction mechanism and a channel de-correlation regularization loss.

CLIP-Hitchhiker~\cite{vtr:clip-hitchhiker} observed that most single-modal temporal modeling and aggregation strategies under-perform a simple average temporal pooling operation. To address this, it introduced a query-guided temporal aggregation baseline, which achieved performances superior to SOTA methods. CenterCLIP~\cite{vtr:centerclip} proposes a more efficient full fine-tuning paradigm to transfer CLIP's knowledge, where token clustering is conducted within each video segment and only center tokens are retained and further processed to compute the video-level embedding. TS2-Net~\cite{vtr:ts2net} combines a temporal token shifting operation and a spatial token selection module to enhance the spatial-temporal video representation learning for better retrieval. CLIP2TV~\cite{vtr:clip2tv} constructs soft labels from the intra-modality similarity matrices to guide the learning of a fusion-based video-text matching module. X-CLIP~\cite{vtr:x-clip} transfers the knowledge of pretrained CLIP models to end-to-end video–text retrieval through multi-grained contrastive learning across video–sentence and frame–word representations. In addition, an Attention over Similarity Matrix (AOSM) mechanism is employed to integrate similarity scores across multiple cross-modal levels. CLIP-ViP~\cite{vtr:clip-vip} introduces video proxy tokens as intermediaries for local–global information exchange, and further conducts video–text post-pretraining followed by downstream fine-tuning. To mitigate the injection of irrelevant visual information during video representation learning, X-pool~\cite{vtr:x-pool} introduces a text-conditioned frame embedding pooling module, enabling more effective transfer of CLIP. VOP~\cite{vtr:vop} introduces a cooperative text–video prompt-tuning strategy to mitigate catastrophic forgetting of pretrained vision–language knowledge while substantially reducing the number of trainable parameters. Considering the mutual dependency between contrastive dual encoders and the spatio-temporal characteristics of video data, VOP inserts learnable prompt tokens at the inputs of each text and vision encoder block, with positional, contextual, and functional aspects of video prompts explicitly modeled to enable effective transfer. 

CLIPPING~\cite{vtr:clipping} proposes an efficient knowledge distillation framework with a novel Student-As-Base (SAB) scheme, which regards the student's intermediate features as the bases of the teacher's feature space to absorb the pretrained knowledge. PromptSwitch~\cite{vtr:ptsw} inserts a cube tensor as learnable prompts into the CLIP image encoder, where the prompt cube can be flexibly transposed to assist in the peer-to-peer communication between each frame pairs in the video. RAP~\cite{vtr:rap} designs a low-rank modulation operation in the adapter module to inject the temporal sparsity and correlation characteristics for efficient image-to-video transfer. MV-Adapter~\cite{vtr:mv-adapter} equips the adapter module with two novel components of temporal adaptation and cross-modality typing that can effectively tackle the task of parameter-efficient video-text retrieval. TeachCLIP~\cite{vtr:teachclip} introduces a multi-grained knowledge distillation framework to perform fine-grained frame-level distillation and coarse-grained video-level distillation, leading to strong performance in video–text retrieval. DiscoVLA~\cite{vtr:discovla} introduces parameter-efficient fusion and distillation modules to simultaneously mitigate the vision, language, and alignment gaps in existing image-to-video transfer learning paradigms.

\subsection{Video Action Recognition}
Video action recognition is a classical task in video understanding, which aims to recognize the category of action being performed in a trimmed video. Over the years, there have been plentiful research works making efforts to transfer the powerful pretrained ILFMs for tackling video action recognition. ActionCLIP~\cite{var:actionclip} is a pioneering work that adopts a ``pretrain, prompt, and fine-tune" paradigm for transferring the knowledge of CLIP, which formulates action recognition into a multimodal learning problem by extracting text representations of class labels for video-text matching. X-CLIP~\cite{var:x-clip} introduces a new framework to expand existing large-scale image-text pretrained models for general video recognition, where a cross-frame communication attention module and a video-specific textual prompting strategy are incorporated. EVL~\cite{var:evl} proposes to utilize a sideway transformer decoder to progressively aggregate multiple intermediate features from a frozen pretrained backbone to a learnable query token that is regarded as the global feature for predicting action classes. ST-Adapter~\cite{var:st-adapter} presents an efficient spatio-temporal adapter which is instantiated as a 3D convolution module inserted at the front end of each pretrained block in cascade.

Adaptformer~\cite{var:adaptformer} designs another kind of adapter module which is inserted as a parallel learnable branch for merging into the original MLP's output. AIM~\cite{var:aim} jointly incorporates cascaded adapters into the spatial and temporal attention layers and adds an adapter in parallel to the MLP layer to achieve efficient transfer learning. BIKE~\cite{var:bike} explores a bidirectional cross-modal knowledge transfer approach, where the attributes-category association and video concept spotting mechanisms are introduced to complement the unidirectional video-to-text matching and enhance the temporal awareness of video representations. ViFi-CLIP~\cite{var:ViFi-CLIP} shows that simple full fine-tuning of pretrained CLIP models with late temporal pooling generalizes well to supervised settings and a vision-language prompting scheme works well in low-data regimes. Text4Vis~\cite{var:text4vis} demonstrates that a new paradigm with frozen classifiers initialized from pretrained textual embeddings can achieve more efficient transfer learning as compared to previous contrastive-based paradigms. 

DiST~\cite{var:dist} disentangles spatial and temporal modeling to the pretrained backbone and a learnable network, and further utilizes a side branch to integrate multiple intermediate features from the spatial and temporal streams. UniFormerV2~\cite{var:uniformerv2} designs a sophisticated transfer method by additionally learning local temporal modeling and global spatio-temporal modeling on top of the pretrained global spatial modeling and a multi-stage fusion block is utilized to aggregate video tokens from different semantic levels. DUALPATH~\cite{var:dualpath} explicitly separates the image-to-video adaptation into a spatial path with parallel adapters and a temporal path with serial adapters followed by a late path fusion for action classification. In addition, a new manner to imitate spatial modeling to achieve efficient temporal modeling via token interactions in a grid-like frameset is also proposed. MoMa~\cite{var:moma} proposes a new temporal modeling method called sequence modulation operation, which is achieved by efficient trainable mamba layers to inject fine-grained spatio-temporal information into intermediate features. D$^2$ST~\cite{var:d2st} develops a dual-pathway adapter that enables disentangled adaptation of spatial and temporal features, while maintaining full compatibility with pretrained single-stream image-text models.

\subsection{Video Captioning}
Video captioning aims to generate natural language descriptions of video content. It has evolved from early template-based methods~\cite{caption:recon,caption:multi,caption:dense} to the adaptation of large pre-trained foundation models. Many works adopt a post-network tuning strategy to enhance features from frozen image encoders. Clip4Caption~\cite{caption:clip4caption} introduces a post-network for temporal segment sampling and utilizes contrastive pretraining to learn text-aligned video embeddings from CLIP features. EvCap~\cite{caption:evcap} also employs a post-network approach to fuse global visual contexts with specific action and object cues extracted by a pretrained CLIP model. mPLUG-2~\cite{caption:mplug2} introduces a partial pretraining paradigm where a dual-vision encoder decouples spatial and temporal modeling by augmenting shared spatial self-attention layers with a novel local temporal modeling module for video inputs. MAMS~\cite{caption:mams} introduces a model-agnostic Side-Tuning framework that utilizes a module selector to adaptively choose between captioning modules of different sizes and incorporates an adaptive attention masking scheme to focus on important visual tokens. 

Similarly, CM$^2$~\cite{caption:remember} enhances CLIP features using multi-scale temporal convolutions and a cross-modal memory module that retrieves relevant textual semantics as external cues. In contrast, SWINBERT~\cite{caption:swinbert} proposes an end-to-end fine-tunable architecture, employing a Video Swin Transformer to extract dense spatio-temporal tokens which are then processed by a multimodal transformer with sparse attention for effective long-range modeling. More recent approaches focus on improving temporal precision. Vid2Seq~\cite{caption:vid2seq} presents a unified framework for joint event captioning and localization by augmenting the language model's vocabulary with special time tokens that serve as an explicit temporal modeling component. VTG-LLM~\cite{caption:vtgllm} enhances pretrained encoders by injecting temporal cues like sequential and absolute time embeddings and employs a slot-based token compression mechanism for the efficient handling of long video sequences. HierarQ~\cite{hierarq} leverages a trainable hierarchical Q-Former as post network, equipped with a two-stream feature modulator and dedicated memory banks, to compress frozen video features into a compact representation for a LoRA-adapted LLM.

\begin{table*}[ht!]
  \mbox{}\hfill
  \begin{minipage}[]{\textwidth}
  \centering
  \footnotesize
  \renewcommand{\arraystretch}{1.0}
  \caption{Experimental results of representative methods in fine-grained video understanding tasks.}
  \setlength{\tabcolsep}{1.2mm}{
 \begin{tabular}{c|c|c|c|c|c|c|c|c}
      \toprule
      \rowcolor{blue!10}
      Task &Category& Method &Venue& ILFM& TETA$\uparrow$&LocA$\uparrow$ &AssocA$\uparrow$ & ClsA$\uparrow$  \\
      \midrule
      \multirow{6}{*}{\rotatebox{90}{
      \begin{minipage}{1.5cm}
      \centering
      OV-MOT
      \end{minipage}
      }}
      &Knowledge Distillation&OVTrack \cite{OVMOT:OVTrack} & CVPR 2023&CLIP&35.5&49.3&36.9&20.2 \\
      &F.Tuning with Adapter&MASA \cite{OVMOT:MASA}& CVPR 2024 &G-DINO& 36.9&55.1&36.4&19.3\\
      &F.Tuning with Extra Models&NetTrack \cite{OVMOT:NetTrack} & CVPR 2024& G-DINO&33.0&45.7&28.6&\textbf{24.8}\\
      &Post Network Tuning&OVTR \cite{OVMOT:OVTR} & ICLR 2025 & CLIP &  36.6&52.2&37.6&20.1\\
      &Knowledge Distillation&GLATrack\cite{OVMOT:GLATrack}&ACM MM 2024&CLIP&37.9&54.3&38.7&20.8\\
      &Knowledge Distillation&OVSORT\cite{OVMOT:OVSORT}&TMM 2025&CLIP&\textbf{38.2}&\textbf{55.3}&\textbf{39.9}&19.4\\
      
      \rowcolor{blue!10}
      \toprule
      Task &Category& Method &Publication&ILFM&m\_vIoU $\uparrow$ & vIoU@0.3 $\uparrow$& vIoU@0.5 $\uparrow$& m\_tIoU $\uparrow$  \\
      \midrule
      \multirow{8}{*}{\rotatebox{90}{
      \begin{minipage}{1.5cm}
       \centering
      STVG
      \end{minipage}
      }}
      &Full Fine-Tuning&STVGBert\cite{stvg:STVGBert}&ICCV 2021&ViLBERT & 24.0 & 30.9 & 18.4& -  \\
      &Full Fine-Tuning&TubeDETR\cite{stvg:TubeDETR} & CVPR 2022 &MDETR  & 30.4 & 42.5 & 28.2& 48.1   \\
      &Full Fine-Tuning&STCAT\cite{stvg:STCAT} & NeurIPS 2022&MDETR &{33.1} &{46.2} & 
     {32.6} &50.8   \\
        &F.Tuning with Extra Models& CoSD\cite{stvg:CoSD} &CVPR 2023&MDETR  &33.7 &  47.2 &  32.8 &50.0  \\
        &F.Tuning with Extra Models& CoSTA\cite{stvg:CoSTA} &AAAI 2024&MDETR  &\textbf{35.1}&\textbf{48.4}&\textbf{34.0} &\textbf{52.1}  \\
        &F.Tuning with Extra Models& CGSTVG\cite{stvg:CGSTVG} &CVPR 2024&MDETR  &34.0&47.7&33.1 & 51.4  \\
        &F.Tuning with Extra Models& TASTVG\cite{stvg:TASTVG} &ICLR 2025&MDETR  & 34.4&48.2&33.5 &51.7 \\
        &Post-Network Tuning&SpaceVLLM\cite{stvg:spacevllm}&Arxiv 2025&SigLIP&27.4&39.1&26.2&47.7\\
        \toprule
        \rowcolor{blue!10}
      Task &Category& Method &Publication&ILFM&R@0.3 $\uparrow$ & R@0.5 $\uparrow$& R@0.7 $\uparrow$& m\_IoU $\uparrow$  \\
      \midrule
      \multirow{11}{*}{\rotatebox{90}{
      \begin{minipage}{1.5cm}
       \centering
      TVG
      \end{minipage}
      }} 
      &Post Network Tuning&MomentDETR\cite{tvg:qvhighlight} & NeurIPS 2021 &CLIP  & 65.8 & 52.1 & 30.6& 45.5   \\
      &F.Tuning with Extra Models&VDI\cite{tvg:vdi}&CVPR 2023&CLIP & - & 52.3 & 31.4& -  \\
      &F.Tuning with Adapter& LlaViLo\cite{tvg:llavilo} &ICCV 2023&LLaVA&  - &  55.7 &  33.4 &-  \\
      &Post Network Tuning&UniVTG\cite{tvg:univtg} & ICCV 2023 &CLIP  & 70.8 & 58.0 & 35.7& 50.1   \\
      &Side-Tuning&R$^2$-tuning\cite{tvg:r2tuning} & ECCV 2024&CLIP &70.9 &59.8 & 37.0 &50.9   \\
        &F.Tuning with Extra Models& Vtimellm\cite{tvg:vtimellm} &CVPR 2024&LLaVA  &51.0&27.5&11.4 & 29.3  \\
        &F.Tuning with Extra Models& VTG-LLM\cite{caption:vtgllm} &AAAI 2025&LLaVA  &52.0&33.8&15.7 &-   \\
        &F.Tuning with Extra Models& Seq2Time\cite{tvg:seq2time} &CVPR 2025&LLaVA  & -&31.2&13.7 &- \\
        &F.Tuning with Extra Models& NumPro\cite{tvg:numpro} &CVPR 2025&LLaVA  & 63.8&42.0&20.6 &41.4 \\
        &F.Tuning with Extra Models& UniTime\cite{tvg:unitime} &Arxiv 2025&Qwen2 VL  & -&- &31.9 &52.2 \\
        &Post Network Tuning& VideoTG-R1\cite{VTG:VideoTGR1} &Arxiv 2025&Qwen2.5 VL  & \textbf{79.5} &\textbf{64.8}&\textbf{39.2} &\textbf{55.9} \\
      \bottomrule
      \end{tabular}}
    \label{tab:fine-grained-transfer}
  \end{minipage}
  ~\hfill
  \end{table*}

\subsection{Video Question Answering}

Pioneering methods~\cite{vqa:tgif, vqa:motion, vqa:3d, vqa:graph} either apply hierarchical spatio-temporal modeling strategies, employing two-stage architectures to first extract key frame clips and then perform reasoning~\cite{vqa:hierarchical}, or construct spatio-temporal relational graphs using graph neural networks. Recently, more and more methods~\cite{vqa:atp,vqa:temadapter} consider addressing the VideoQA problem utilizing the pre-trained foundation models.

Many studies have focused on exploring frozen-feature strategies, such as knowledge distillation and post-network tuning, for image-to-video transfer learning. VideoDistill~\cite{vqa:videodistill} introduces a language-aware gating module that dynamically selects and refines frame-level visual features. This design facilitates goal-oriented selective knowledge distillation and effectively mitigates language bias that often arises when adapting image-based models to video data.
BIMBA~\cite{vqa:bimba} adopts the post-network tuning strategy by introducing a Mamba-based token compression module on top of pre-trained foundation models. The module models long-range spatiotemporal dependencies and compresses redundant tokens, thereby improving efficiency. LinVT~\cite{vqa:linvt} implements post-network tuning through a plug-and-play Linear Video Tokenizer module. Placed after a frozen visual encoder, this post-network employs a Spatio-Temporal Visual Token Refiner (SVR) and a Text-Conditioned Token Aggregator (TTA) to produce a compact, question-relevant video representation for the subsequent large language model. HierarQ~\cite{hierarq} develops a post-network tuning module in the form of a trainable hierarchical Q-Former equipped with task-aware modulators and memory mechanisms. It generates a compact yet informative spatiotemporal representation, which is subsequently fed into a frozen large language model that is further adapted through a lightweight adapter for textual answer generation. TEA~\cite{Zhang_Zeng_Shen_Wu_Zhou_Ma_2025} extends TextVQA~\cite{singh2019towards} from image to video by incorporating a spatio-temporal recovery module, which includes a temporal convolution block and an OCR-enhanced relative spatial bias, into pretrained T5 language model. 

Recent progress in MLLMs has established a new paradigm for Video QA, heavily reliant on the integration of adapted visual and textual features. Atemporal Probe (ATP) \cite{vqa:atp} introduces a fine-tuning with prompt tuning approach that selects a single representative key frame to serve as a learnable visual prompt for a frozen vision encoder. Q-ViD \cite{vqa:qvid} leverages a pre-trained InstructBLIP model \cite{vqa:instruct} and incorporates tailored instructional prompts to generate descriptive video captions, using them as contextual cues for improved question answering. Similarly, ViTiS \cite{vqa:zero} employs multimodal prompt learning together with a visual mapping network to inject task-specific cues into image-language models, enabling strong performance with only few additional learnable parameters and limited training data. Other works explore adapter-based fine-tuning. Tem-Adapter \cite{vqa:temadapter} introduces a visual temporal aligner to predict future video states and a textual semantic aligner to refine embeddings for cross-modal alignment. TR-Adapter~\cite{vqa:tradapter} proposes a temporal reasoning adapter designed to capture temporal relationships and facilitate knowledge transfer from language models. Unlike prior methods that maintain the original backbone architecture and perform adaptation through external lightweight modules, TC-LLaVA \cite{vqa:vcllava} modifies the attention computation in foundation models through partial tuning. Specifically, it introduces a Temporal-Aware Dual RoPE to encode temporal positional information and a Frame-wise Block Causal Attention Mask to enhance interactions among visual tokens.

\section{General Video Understanding}
Unlike most existing image-to-video transfer learning methods, which are tailored for one specific task, there have been some methods that aim to transfer the pretrained image-text knowledge across multiple tasks to achieve general video understanding as a video-language foundation model. 

FitCLIP~\cite{gvu:fitclip} proposes to train a student model by a small number of labeled video-text pairs and a large number of video-text pseudo labels, where the student model is then fused with the teacher model by weight ensembling for zero-shot video understanding. STAN~\cite{gvu:stan} presents a spatial-temporal auxiliary network as a side-tuning approach to perform temporal interaction and feature merging in the additional side branch, which achieves SOTA performances across video-text retrieval and video action recognition.  Video-ChatGPT~\cite{general:videochatgpt} employs a frozen CLIP-L/14 visual encoder to extract both spatial and temporal video features by averaging frame-level representations across their respective dimensions. The resulting spatiotemporal features are subsequently projected into the large language model's input space via a learnable linear layer through video instruction tuning. Video-LLaMA~\cite{general:videollama} incorporates an additional Video Q-Former with temporal position encoding and a frozen LLM as the post-network following a pretrained frozen BLIP-2~\cite{intro:bilp2} visual encoder. The architecture also introduces a similarly structured audio branch, enabling the model to process both visual and auditory modalities simultaneously.  Video-LLaVA~\cite{general:videollava} utilizes LanguageBind~\cite{general:languagebind}, initializing from Open-CLIP as a frozen video encoder, and subsequently fine-tunes the LLM and projection layer following the methodology established in LLaVA\cite{llava}. UniformV2~\cite{intro:uniformerv2} demonstrates that incorporating specifically designed local and global relation aggregators after a large-scale image-text pre-trained visual encoder, followed by video pre-training, can significantly enhance performance.  Based on CLIP's multimodal contrastive learning with video data, InternVideo~\cite{general:internvideo} incorporates an additional masked video encoder. This encoder learns meaningful spatiotemporal representations by reconstructing masked video tokens. Furthermore, the framework unifies their representations through lightweight interactive model supervision. Experiments confirm that features learned via generative and contrastive training are complementary, and their integration yields superior performance compared to training either objective independently. InternVideo2~\cite{general:internvideo2} employs InternVL~\cite{intro:internvl} as the teacher model, utilizing a distillation paradigm to transfer knowledge from the ILFMs to the video encoder before proceeding with subsequent multimodal alignment and LLM-linking. InternVideo2.5~\cite{general:internvideo2.5} employs InternVL2.5~\cite{intro:internvl2.5} as its visual encoder and introduces hierarchical token compression alongside task preference optimization during pre-training, enabling robust long and rich context understanding in MLLMs.

\begin{table*}[ht!]
  \mbox{}\hfill
  \begin{minipage}[]{\textwidth}
  \centering
  \footnotesize
  \renewcommand{\arraystretch}{1.0}
  \caption{Experimental results of representative methods in coarse-grained video understanding tasks.}
  \setlength{\tabcolsep}{1.5mm}{
 \begin{tabular}{c|c|c|c|c|c|c|c|c}
      \toprule
       \rowcolor{blue!10}
        Task &Category& Method &Venue& ILFM&R@1$\uparrow$ & R@5$\uparrow$& R@10$\uparrow$& R@sum$\uparrow$  \\
      \midrule
      \multirow{5}{*}{\rotatebox{90}{
      \begin{minipage}{1.5cm}
       \centering
      VTR
      \end{minipage}
      }} 
      &Full Fine-Tuning&CLIP4Clip~\cite{vtr:clip4clip} & Neurocomp. 2022 &CLIP & 44.5 & 71.4 & 81.6 & - \\
      &Prompt Tuning&VOP~\cite{vtr:vop}&CVPR 2023&CLIP & 44.6 & 69.9 & 80.3 & - \\
      &F.Tuning with Adapter&MV-Adapter~\cite{vtr:mv-adapter} &CVPR 2024&CLIP& 46.2 & 73.2 & 82.7 & -  \\
      &Knowledge Distillation&TeachCLIP~\cite{vtr:teachclip} & CVPR 2024 &CLIP & 46.8 & 74.3 & - & 203.7  \\
      &F.Tuning with LoRA&DiscoVLA~\cite{vtr:discovla} & CVPR 2025&CLIP &\textbf{50.5} &\textbf{75.6} & \textbf{83.8} &\textbf{209.9}  \\
      \toprule
      \rowcolor{blue!10}
    Task &Category& Method &Publication& ILFM&K400 Top-1$\uparrow$ & K400 Top-5$\uparrow$& SSv2 Top-1$\uparrow$& SSv2 Top-5$\uparrow$  \\
      \midrule
      \multirow{6}{*}{\rotatebox{90}{
      \begin{minipage}{1.5cm}
       \centering
      VAR
      \end{minipage}
      }}
      &Side-Tuning&EVL~\cite{var:evl} & ECCV 2022 &CLIP & 87.3 & - & 66.7 & - \\
      &F.Tuning with Adapter&AIM~\cite{var:aim}&ICLR 2023&CLIP & 87.5 & 97.7 & 70.6 & 92.7 \\
      &F.Tuning with Adapter&DUALPATH~\cite{var:dualpath}&CVPR 2023&CLIP & 87.7 & 97.8 & 71.4 & 93.4 \\
      &Side-Tuning&DiST~\cite{var:dist} &ICCV 2023&CLIP& \textbf{88.0 }& 97.9 & 73.1 & 93.2 \\
      &F.Tuning with Adapter&ZeroI2V~\cite{var:zeroi2v} & ECCV 2024 &CLIP & 87.2 & 97.6 & 72.2 & 93.0  \\
      &F.Tuning with Adapter&MoMa~\cite{var:moma} & ICML 2025&CLIP &87.8 &\textbf{98.0} & \textbf{73.8} &\textbf{93.6}  \\
      \toprule
      \rowcolor{blue!10}
    Task &Category& Method &Publication& ILFM&MSRVTT$\uparrow$ & NExT-QA$\uparrow$& MSVD-QA$\uparrow$& How2QA$\uparrow$  \\
      \midrule
      \multirow{8}{*}{\rotatebox{90}{
      \begin{minipage}{1.5cm}
       \centering
      VideoQA
      \end{minipage}
      }} 
      &Prompt tuning&ATP\cite{vqa:atp} & CVPR 2022 &CLIP  & 93.2 & 54.3 & -& 65.1   \\
      &F.Tuning with Adapter&Tem-Adapter\cite{vqa:temadapter}&ICCV 2023&CLIP & 94.3 & - & -& -  \\
      &Prompt tuning& ViTiS\cite{vqa:zero} &ICCV 2023&CLIP&  - &  - &  47.8 &- \\
      &Prompt tuning& Q-Vid\cite{vqa:qvid} &ACL 2024&CLIP&  - &  66.3 &  - &\textbf{71.4} \\
      &Knowledge Distillation&VideoDistill\cite{vqa:videodistill} & CVPR 2024 &CLIP  & \textbf{97.8} & - & \textbf{49.2}& -   \\
      &Post Network Tuning&BIMBA\cite{vqa:bimba} & CVPR 2025&LLaVA &- &83.7 & - &-  \\
      &Post Network Tuning&LeAdQA\cite{vqa:leadqa} & Arxiv 2025&Qwen2.5VL &- &80.6 & - &-  \\
      &Post Network Tuning&SG-VLM\cite{vqa:sgvlm} & Arxiv 2025&InternVL &- &\textbf{85.1} & - &-  \\
      \toprule
      \rowcolor{blue!10}
    Task &Category& Method &Publication& ILFM&BLEU4$\uparrow$ & ROUGE-L$\uparrow$& METEOR$\uparrow$& CIDEr$\uparrow$  \\
      \midrule
      \multirow{8}{*}{\rotatebox{90}{
      \begin{minipage}{2cm}
      \centering
      Captioning
      \end{minipage}
      }}
      &Post Network Tuning&CLIP4Caption\cite{caption:clip4caption} & ACM MM 2021 &CLIP  & 47.2 & 64.8  & 31.2& 60.0   \\
      &Full Fine-Tuning&SWINBERT\cite{caption:swinbert}&CVPR 2022&CLIP & 45.4 & 64.1 & 30.6& 55.9 \\
      &Partial tuning& mPLUG-2\cite{caption:mplug2} & ICML 2023&CLIP &57.8 &70.1 & 34.9 &80.3  \\
      &F.Tuning with Extra Models& Vid2Seq\cite{caption:vid2seq} &CVPR 2023&CLIP&  - &  - &  30.8 &64.6 \\
      &Side-Tuning&MAMS\cite{caption:mams} & AAAI 2025 &CLIP  & \textbf{60.0} & \textbf{71.2} & 34.7& \textbf{82.9}   \\
      &Post Network Tuning&HierarQ\cite{hierarq} & CVPR 2025& LLaVA &- &- & \textbf{35.1} &80.5   \\
      &Post Network Tuning&OHRC\cite{caption:ohrc} & TCSVT 2025& CLIP &46.9 &67.3 & 33.4 &59.0   \\
      &F.Tuning with Adapter&Q-Adapter\cite{caption:qadapter} & ACM MM 2025& Qwen2.5VL &49.8 &62.7 & 32.3 &67.4  \\
      \bottomrule
      \end{tabular}}
    \label{tab:coarse-grained-transfer}
  \end{minipage}
  ~\hfill
  \end{table*}
  
\section{Experimental Analysis}
In the following, we present the experimental results of various transfer learning techniques applied to different video-text understanding tasks, followed by a detailed analysis.

\subsection{Open Vocabulary Multi-Object Tracking}

\noindent \textbf{Datasets and Metrics.} For the OV-MOT task, we use TAO\cite{OVMOT:TAO} as the benchmark to compare existing image-to-video transfer methods.  TAO is an open-world video dataset with detailed annotations, supporting tracking of arbitrary objects. This set is divided into three subsets: training, validation, and testing, comprising 500, 988, and 1,419 videos, respectively. The OV-MOT method is primarily evaluated using the TETA metric. TETA\cite{OVMOT:TETA}, an extension of HOTA\cite{OVMOT:HOTA}, desiged for long-tail scenarios. TETA comprises three components: the localization score (LocA), which measures bounding box alignment; the classification score (ClsA), which evaluates object category correctness; and the association score (AssocA), which assesses ID consistency across frames. The overall TETA score is the arithmetic mean of these three sub-scores.

\noindent \textbf{Results and analysis.} In the OV-MOT task, researchers explored multiple image-to-video transfer paradigms, including distillation, adapter modules, and post-network tuning, while also experimenting with two distinct foundation models: CLIP\cite{intro:clip} and G-DINO \cite{intro:groundingdino}. As shown in Table \ref{tab:fine-grained-transfer}, no single method clearly outperforms the others or achieves superior performance across all metrics. Specifically, the CLIP-based GLATrack model \cite{OVMOT:GLATrack}, which performs image-to-video transfer via knowledge distillation, achieves the highest scores on TETA and AssocA, surpassing other competitors by more than 1\%. In contrast, the G-DINO-based approaches MASA \cite{OVMOT:MASA} and NetTrack \cite{OVMOT:NetTrack} perform better on LocA and ClsA, respectively, highlighting that different methods excel in different evaluation criteria.

\subsection{Spatio-Temporal Video Grounding}

\noindent \textbf{Datasets and Metrics.} For the STVG task, VidSTG~\cite{stvg:VidSTG} is currently the most widely-used benchmark dataset. It contains 99,943 video-text 
pairs, including 44,808 declarative sentences and 55,135 interrogative sentences from 10,303 videos spanning 80 
object categories. For clearer comparison, we evaluate different methods under the declarative sentence 
setting. Following \cite {stvg:CGSTVG, stvg:TASTVG}  we use mean vIoU as the main metric. vIoU is computed as: $vIoU = \frac{1}
{\left|T_u\right|}\sum_{t\in T_i}IoU(\hat{b}_t, b_t),$
where $T_i$ and $T_u$ denote the intersection and union of the annotated and predicted time intervals, and $\hat{b}_t, b_t$ are the predicted and ground truth bounding boxes for the $t$-th frame, respectively. 
We also report vIoU@R which indicates the proportion of samples with vIoU higher than $R$. Following \cite{stvg:TubeDETR}, we report tIoU evaluating the temporal grounding performances. It is defined as: $tIoU=\frac{\left|T_i\right|}{\left|T_u\right|}.$ 

\noindent \textbf{Results and analysis.} The comparative results on VidSTG-Declarative\cite{stvg:VidSTG} are shown in Table II. Most fully-supervised STVG methods adopt MDETR as the foundation model for its visual grounding capability. However, relying solely on its image-level features limits temporal localization accuracy. Recent methods therefore integrate auxiliary models (e.g., SlowFast\cite{slowfast}) to extract video-level features, improving temporal performance—tIoU rose from 48.1 (TubeDETR\cite{stvg:TubeDETR}) to 52.1 (CoSTA\cite{stvg:CoSTA}). This indicates that combining both feature types enables more precise spatio-temporal localization, though at increased computational cost (e.g., TASTVG\cite{stvg:TASTVG} required 32 A100 GPUs for 15 hours).

\subsection{Temporal Video Grounding}
\noindent \textbf{Datasets and Metrics.}
To address the TVG problem, numerous datasets have been constructed with video–text pairs and detailed annotations specifying the temporal spans of events described in the text. Specifically, TACoS~\cite{tvg:tacos}, Charades-STA~\cite{tvg:tall} and ActivityNet Captions~\cite{tvg:dense} are the three most widely used benchmarks in the community. Following~\cite{tvg:tall}, most existing approaches are evaluated in Charades-STA~\cite{tvg:tall} and report their performance in terms of metrics ``Rn@m" and ``mIoU". Metric ``Rn@m" is employed to compute the percentage of samples where the top-$n$ predicted segments have an IoU greater than a predefined threshold $m$, where $m$ is typically defined as $\{0.1,0.3,0.5,0.7\}$. Metric ``mIoU" calculates the mean IoU between the predicted segments and ground truth on all testing data.

\noindent
\textbf{Results and Analysis.} 
We report comparison results on the Charades-STA~\cite{tvg:tall} set with ``Rn@m" and ``mIoU" metrics. As shown in Table \ref{tab:fine-grained-transfer}, the results for the TVG task reveal a clear performance gap between different foundation models. Previously, the CLIP-based method R$^2$-tuning \cite{tvg:r2tuning} achieved the best performance with 50.9\% on mIoU, surpassing LLaVA-based methods like NumPro\cite{tvg:numpro} with 41.4\% mIoU. It is worth noting that NumPro relies on a complex Fine-Tuning with Extra Models strategy, while R$^2$-tuning uses Side-Tuning based on MLLM. However, the recent VideoTG-R1\cite{VTG:VideoTGR1} built on Qwen2.5-VL obtains the highest scores across all metrics, reaching 55.9\% in the term of mIoU. Surprisingly, VideoTG-R1 employs a simple Post Network Tuning strategy but outperforms all previous methods. This demonstrates that a stronger foundation model can achieve superior temporal localization without complex auxiliary modules.

\subsection{Video-Text Retrieval}

\noindent \textbf{Datasets and Metrics.} 
To facilitate VTR research, several datasets pairing videos with text descriptions have been created to evaluate cross-modal alignment. Among these, MSR-VTT~\cite{vqa:msrvtt} is a widely used benchmark. Following common practice, most methods are evaluated on this dataset and report R@1, R@5, R@10, and R@sum. R@1, R@5, and R@10 measure recall among the top 1, 5, and 10 results, reflecting ranking accuracy. R@sum aggregates these three scores to summarize overall retrieval performance.

\noindent \textbf{Results and analysis.} By examining the results on MSR-VTT~\cite{vqa:msrvtt} in Table~\ref{tab:coarse-grained-transfer} for the task of Video-Text Retrieval, it can be observed that most recent methods are built upon the pretrained CLIP model due to its strong visual-textual correspondence capability. Among these, fine-tuning with LoRA achieves the best performance, significantly surpassing other feature-modification transfer learning strategies such as prompt tuning, knowledge distillation, and adapter-based tuning, in the metric of R@1. As expected, directly fusing CLIP-extracted features with a post-network tuning scheme performs worse than most approaches that update CLIP parameters, highlighting the importance of adapting the original foundation model to the target task and dataset for optimal performance.

\subsection{Video Action Recognition}

\noindent \textbf{Datasets and Metrics.}
To support research in VAR, several influential datasets have been developed. Foundational benchmarks include UCF101~\cite{dataset:ucf101} and HMDB51~\cite{dataset:hmdb51}, which established early standards for the task. More recently, the large-scale Kinetics series Kinetics-400~\cite{dataset:kinetics400} and Something-Something V2~\cite{dataset:ssv2} have become the most widely adopted benchmarks for evaluating VAR models. Following common practice, model performance on these datasets is primarily evaluated using Top-1 and Top-5 accuracy. Top-1 accuracy measures the percentage of videos for which the model's single highest-probability prediction is correct. Top-5 accuracy measures the percentage of videos where the ground-truth label is among the model's five most confident predictions, which is particularly informative for datasets with a large number of classes.

\noindent \textbf{Results and analysis.} Table \ref{tab:coarse-grained-transfer} (termed as VAR) reports a detailed comparison of several representative image-to-video transfer methods for video action recognition based on the CLIP model. Overall, finetuning with Adapter strategy achieves better results than side-tuning strategy in this task. MoMa \cite{var:moma} obtains the best results on both K400 and SSv2 sets, which employs a mamba architecture to learn long-term fine-grained spatio-temporal information. The comparison between EVL \cite{var:evl} and DiST \cite{var:dist} demonstrates the effectiveness of disentangling the spatial and temporal modeling in image-to-video transfer learning.

\subsection{Video Question Answering}

\noindent \textbf{Datasets and Metrics.} VideoQA is typically evaluated using metrics such as Top-1 Accuracy and Top-K Accuracy. The VideoQA task typically comprises three categories: \textit{i) Multiple-Choice:} Requires selecting the correct answer from a predefined set. Representative benchmarks such as MSRVTT-MC \cite{vqa:msrvtt}. \textit{ii) Open-Ended:} Involves generating free-form responses that often necessitate logical or creative reasoning. Popular datasets include MSRVTT-QA \cite{vqa:msrvtt}, MSVD-QA \cite{vqa:msvd}, and ActivityNet-QA \cite{vqa:act}. \textit{iii) Long-Form:} Demands comprehensive, explanatory answers reflecting deep temporal understanding, with key benchmarks such as How2QA \cite{vqa:how2}. 


\noindent \textbf{Results and analysis.} We present the comparison results on four datasets (MSRVTT-MC\cite{vqa:msrvtt}, NExT-QA\cite{xiao2021next}, MSVD-QA\cite{vqa:msvd} and How2QA\cite{vqa:how2}) with Top-1 Accuracy on Multiple-Choice in Table \ref{tab:coarse-grained-transfer}, where most competitors use CLIP as their backbone to ensure a fair comparison. For standard vision-language models such as CLIP, various transfer methods demonstrate strong and dataset-dependent performance. The distillation-based approach, VideoDistill \cite{vqa:videodistill}, achieves SOTA results on MSRVTT-MC and MSVD-QA with accuracies of 97.8\% and 49.2\%, respectively. In contrast, on the How2QA benchmark, the prompt-tuning method Q-Vid \cite{vqa:qvid} attains the highest score at 71.4\%. These results suggest that for a fixed backbone like CLIP, the most effective adaptation strategy varies across different benchmarks. Conversely, the integration of LLMs substantially elevates performance. On the NExT-QA benchmark, the LLaVA-based BIMBA \cite{vqa:bimba} model, which utilizes a post-network tuning strategy, achieves an accuracy of 83.73\%. This represents a significant improvement over the best CLIP-based method, Q-Vid with an accuracy of 66.3\% on the same dataset. This finding underscores that post-network tuning is a potent transfer strategy for LLM-based models, as it efficiently adapts visual information to unlock the LLM’s powerful semantic reasoning for video understanding.

\subsection{Video Captioning}
\noindent \textbf{Datasets and Metrics.} Prominent open-domain datasets like MSRVTT \cite{vc:MSRVTT} and MSVD \cite{vc:MSVD} assess general descriptive capabilities, while procedural datasets such as YouCook2 \cite{vc:youocook} challenge models with fine-grained, instructional content. Performance on these benchmarks is typically quantified by comparing generated captions against human references using a suite of automated metrics. Specifically, BLEU@N \cite{vc:bleu} and ROUGE-L \cite{vc:rouge} focus on lexical similarity and recall, whereas METEOR \cite{vc:meteor} and CIDEr \cite{vc:cider} provide a more semantically informed evaluation by considering synonyms and consensus, collectively ensuring a holistic assessment of linguistic fluency and content fidelity.

\noindent \textbf{Results and analysis.} As shown in Table \ref{tab:coarse-grained-transfer}, recent advancements in video captioning are driven by adapted transfer strategies that outperform traditional fine-tuning. The side-tuning method MAMS \cite{caption:mams} achieves SOTA performance across multiple key metrics (60.0 BLEU4, 71.2 ROUGE-L, 82.9 CIDEr). This success can be primarily attributed to its architectural design; by training a separate, lightweight module while keeping the foundation model frozen, it effectively learns task-specific temporal dynamics without the risk of catastrophic forgetting. The strong performance of the LLaVA-based HierarQ \cite{hierarq}, which excels by aligning visual features with an LLM's powerful reasoning, further validates this trend. These results indicate that the most effective transfer strategies are those that intelligently integrate new knowledge while preserving the core capabilities of the pre-trained backbone.

\section{FUTURE DIRECTIONS AND DISCUSSION}
In this section, we aim to give some insights on key challenges and potential future research directions related to transferring useful knowledge from pre-trained image-language foundation models for video-text multimodal understanding.

\subsection{Challenges and Future Directions}
A principal challenge in adapting vision-language foundation models to video understanding stems from their inherently specialized architectures. For instance, CLIP is designed for image-text alignment, while MDETR and G-DINO focus on visually-grounded object detection. More recent models, such as LLaVA, InternVL, and QwenVL, have extended these capabilities toward general-purpose visual reasoning and dialogue, yet they remain fundamentally designed for the image domain.

Despite their individual strengths, no single model has yet served as a unified solution that performs robustly and comprehensively across diverse video tasks. As a result, video understanding systems often rely on fragmented pipelines that combine separate foundation models for different sub-tasks. Moreover, how to effectively fine‑tune image‑based models—from established ones like CLIP to more modern multimodal LLMs such as LLaVA and InternVL—on sequential video data remains an open research question. 

As the transfer of ILFMs to video-language tasks continues to evolve, several promising directions emerge. Below, we outline three key trends that are likely to shape future efforts.

\noindent\textbf{Unified Transfer Learning Paradigm for General Video Understanding.} 
Since existing ILFMs excel at specific tasks, their cross-modal capabilities can be transferred to address certain video-language understanding problem, such as MDETR for video grounding. A promising future direction involves developing a unified transfer learning framework to transfer a single ILFM for multiple general video-language understanding tasks simultaneously. This can be achieved by training a unified ILFM capable of performing general image-language understanding, or by prioritizing the design of more universal transfer mechanisms. Potential approaches include exploring prompt-based learning paradigms, where task-specific instructions guide model behavior; incorporating video-oriented reinforcement learning to enhance comprehension; and developing parameter-efficient tuning strategies that facilitate parameter sharing across diverse tasks.

\noindent\textbf{Collaborative Transfer of Multiple Foundation Models.} 
Beyond the development of a unified ILFM, another promising direction lies in the effective integration of multiple pre-trained foundation models, each excelling in a distinct modality or capability, to address complex video-language tasks. Within this paradigm, non-LLM-based models can function as external tools within a video understanding pipeline, generating intermediate outputs such as per-frame spatial bounding boxes from models like MDETR or G-DINO. Concurrently, LLM-based models can serve as central agents, performing reasoning-based inference and orchestrating these tools. Future research should focus on fully leveraging the potential of LLMs in this agentic role while developing methodologies to mitigate the associated computational overhead.

\noindent\textbf{Advanced Efficient Methods for Video Frame Processing.} 
While significant progress has been made, efficiently leveraging ILFMs for video frame processing remains an open challenge. Existing adaptation approaches predominantly rely on uniform frame sampling strategies, which are suboptimal due to the inherent high degree of visual redundancy in video sequences. Consequently, a critical research direction involves designing adaptive frame sampling mechanisms that minimize the number of frames processed by ILFMs without compromising task-critical information. Furthermore, enhancing local alignment between video and text modalities is essential for advancing tasks that demand fine-grained understanding, such as spatiotemporal grounding or long-form video reasoning.

These directions highlight the need for more efficient, integrative, and scalable approaches to bridge image-language models and video-language applications, paving the way for more efficient and diverse video understanding systems.

\subsection{More Discussion}
These future directions mainly target video–language understanding, aiming to make ILFMs more general, collaborative, and efficient under video input streams. Beyond understanding, a parallel and rapidly growing line of research investigates transferring generative ILFMs, such as Stable Diffusion (SD)~\cite{FOUND:SD}, to video generation and editing. For example, VideoCrafter1~\cite{GEN:VCRAFT} further refined spatiotemporal U-Net architectures by partial tuning, often using causal 3D convolutions. Stability AI released Stable Video Diffusion (SVD) \cite{GEN:SVD}, which extends and fully fine-tunes image-based Stable Diffusion models by incorporating temporal layers. Tune-A-Video~\cite{EDIT:TAV} transferred pretrained SD to video tasks through side-tuning spatio-temporal attention modules. 



More recent studies have shifted toward training and fine-tuning video-based foundation models, such as SVD~\cite{GEN:SVD} and WAN~\cite{wan2025}, which have demonstrated superior performance as compared with image-to-video transfer methods built upon image–language foundation models. Readers can refer to ~\cite{he2024llmsmeetmultimodalgeneration, sun2024diffusion, melnik2024video} for a more comprehensive and systematic survey.

\section{CONCLUSION}
This survey presents a comprehensive overview of recent advancements in image-to-video transfer learning within multimodal learning, which, to the best of our knowledge, is the first of its kind. We begin by outlining the necessary background, including fundamental concepts and an introduction to common ILFMs. Subsequently, we summarize nine distinct transfer paradigms employed for image-to-video transfer learning. For each video task, we categorize existing ILFM-based methods according to their technical perspectives and the paradigms used for transferring knowledge from ILFMs. In the experimental section, we detail the evaluation settings and present a fair comparison of results to analyze the performance of different transfer paradigms across various tasks. Finally, we discuss several prevailing challenges and suggest promising future research directions for image-to-video transfer learning.


%




\ifCLASSOPTIONcaptionsoff
  \newpage
\fi



%




\bibliographystyle{ieeetr}
\bibliography{ref}

\end{document}